\documentclass[10pt,twocolumn,letterpaper]{article}
\usepackage[T1]{fontenc}

\usepackage[pagenumbers]{wacv}
\usepackage[accsupp]{axessibility}

\usepackage{newtxtext}
\usepackage{newtxmath}
\usepackage{microtype}
\usepackage{wrapfig}
\usepackage[usenames,dvipsnames]{xcolor}

\usepackage{graphicx}
\usepackage{float}
\usepackage{caption}
\usepackage{multirow}
\usepackage{amsmath}
\usepackage{booktabs}
\usepackage{mathtools}
\usepackage{pifont}
\newcommand{\cmark}{\ding{51}}%
\newcommand{\xmark}{\ding{55}}%
\usepackage{float}

\DeclareMathOperator*{\argmin}{argmin} 
\def\belowcaptionsqueeze{\vspace*{-0.25\baselineskip}}

\usepackage[pagebackref,breaklinks,colorlinks]{hyperref}

\usepackage[capitalize]{cleveref}
\crefname{section}{Sec.}{Secs.}
\Crefname{section}{Section}{Sections}
\Crefname{table}{Table}{Tables}
\crefname{table}{Tab.}{Tabs.}

\usepackage{xspace}
\newcommand{\knn}{$k$-NN\xspace}
\newcommand{\myparagraph}[2][\hspace{0.5em}]{\smallskip\noindent\mbox{\textbf{#2}}#1}

\begin{document}

\title{High-Fidelity Zero-Shot Texture Anomaly Localization Using\\ Feature Correspondence Analysis}

\author{Andrei-Timotei Ardelean\qquad Tim Weyrich\\[0.75ex]
Friedrich-Alexander-Universität Erlangen-Nürnberg\\[0.25ex]
{\tt\small \{timotei.ardelean, tim.weyrich\}@fau.de}
}
\maketitle

\begin{abstract}
   We propose a novel method for Zero-Shot Anomaly Localization on textures. 
   The task refers to identifying abnormal regions in an otherwise homogeneous image. 
   To obtain a high-fidelity localization, we leverage a bijective mapping derived from the 1-dimensional Wasserstein Distance. 
   As opposed to using holistic distances between distributions, the proposed approach allows pinpointing the non-conformity of a pixel in a local context with increased precision. 
   By aggregating the contribution of the pixel to the errors of all nearby patches, we obtain a reliable anomaly score estimate. 
   We validate our solution on several datasets and obtain more than a 40\% reduction in error over the previous state of the art on the MVTec~AD dataset in a zero-shot setting.
   Also see {\footnotesize\texttt{\href{https://reality.tf.fau.de/pub/ardelean2024highfidelity.html}{reality.tf.fau.de/pub/ardelean2024highfidelity.html}}}.
\end{abstract}

\section{Introduction}
\label{sec:intro}

Anomaly Detection (AD) refers to discerning between elements that abide by a standard of normality and those which do not. 
Humans are generally able to perform this distinction without the need for an explicit guideline for the standard of normality simply by comparing them to items that agree to the standard \cite{tao2022deep}. 
Even further, we can often find anomalous regions from visual imagery without previous knowledge of how a certain object or material should look, by simply pinpointing what stands out in a single, isolated sample \cite{lowe2012perceptual}. 
This motivates the search for an automatic system able to perform this task, \ie, zero-shot anomaly localization (ZSAL).

Anomaly detection and localization has a wide range of applications. 
Automatically finding defects during manufacturing, identifying forgeries, detecting situations that require attention in medical imaging, and discovering inaccuracies in industrial machines are just a few of the domains where an anomaly detection system could bring considerable benefits.

The computer vision community has lately shown increased interest in solving the problem of anomaly detection and localization, encouraged by the success of deep learning methods on various tasks~\cite{chai2021deep}.
The primary employed strategy is unsupervised learning, modeling normality from a collection of unblemished items.
This removes the need for labeled anomalous data at training time, which can be difficult to acquire; however, most current systems still require numerous curated (normal) samples~\cite{roth_towards_2022, defard2021padim, zavrtanik2021draem, lee2022cfa, Tien_2023_CVPR, Liu_2023_CVPR}. 
To alleviate this requirement, the more challenging task of few-shot~\cite{rudolph2021same, sheynin2021hierarchical, huang2022registration} and zero-shot~\cite{schwartz_maeday_2022, Jeong_2023_CVPR, aota_zero-shot_2023} AL has recently started to be addressed.

\begin{figure}[t]
  \centering
  \includegraphics[width=0.49\linewidth]{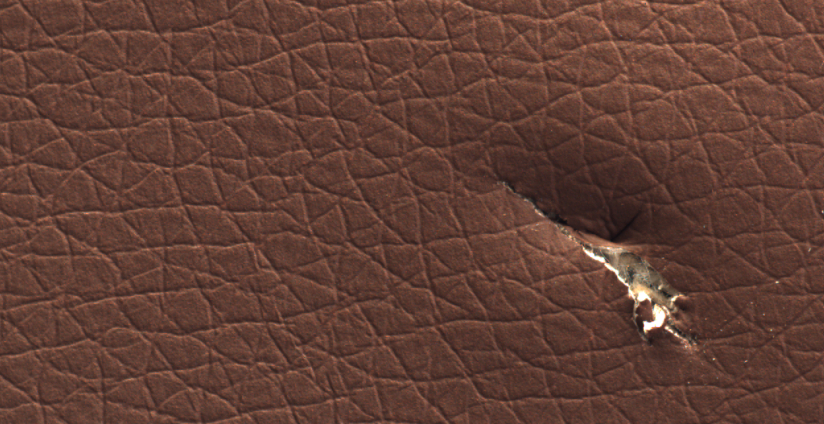}
  \strut\hfill\strut
  \includegraphics[width=0.49\linewidth]{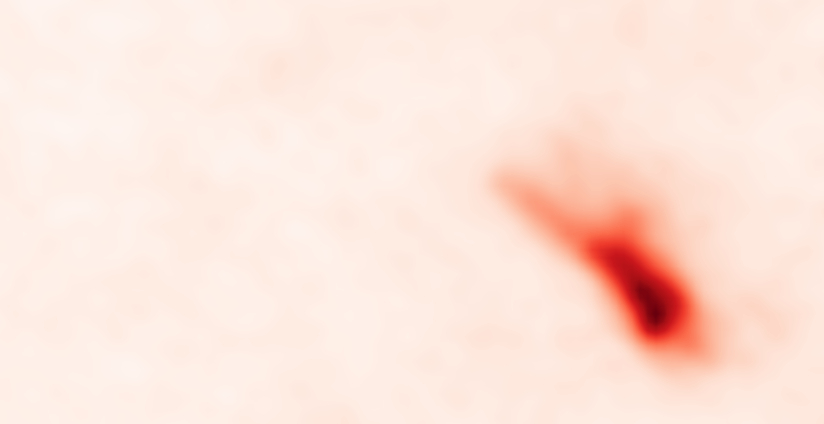}\\[-0.75\baselineskip]
  \caption{\label{fig:teaser}
    Anomaly localization example. 
    \emph{Left:} input texture; 
    \emph{right:} predicted anomaly map.
    \belowcaptionsqueeze}
\end{figure}

We develop a new system designed specifically for anomaly localization that works in a zero-shot setting, identifying the parts that break the homogeneity of a single textured sample (\Cref{fig:teaser}). 
Our main contribution is a novel method for comparing the statistics between different patches in an image or feature map.
To quantify the normality of a pixel location one could trivially compute the average of the nearby features and compare them to a global descriptor, however, as we show, the errors obtained by this method are too coarse for a pixel-level localization of anomalies. 
We analyze different methods for comparing the local statistics of a patch to a (global) reference and show that one can use a bidirectional mapping that implicitly results from the Wasserstein distance to more precisely identify the offending pixels. 
This insight is the core of our Feature Correspondence Analysis (FCA).

\section{Related Work}

The problem of anomaly detection can be posed for various types of data such as weather records \cite{wibisono2021multivariate}, stock market and financial transactions \cite{ahmed2016survey}, acoustic monitoring \cite{cooper2020anomaly}, video surveillance \cite{kaur2018overview}, medical imaging \cite{fernando2021deep}, manufacturing inspection \cite{hsieh2019unsupervised, liu_deep_2023}, etc. 
In this work, we address the detection of anomalies in images, more exactly detecting anomalous regions in otherwise homogeneous or stationary textures. 
This can be formulated as a multi-class segmentation and classification of anomalous pixels \cite{bibi2021edge, oliveira2019lightweight}, or in a simpler setting, as a binary separation between normal and anomalous regions~\cite{cohen2020sub, defard2021padim, li2021cutpaste, rudolph2021same, Yi_2020_ACCV, zavrtanik2021draem}.
We focus on the latter, usually referred to as anomaly localization (AL) despite dealing with pixel-level segmentation (as opposed to localization understood in the context of object detection). 
AL can be considered a superclass of the anomaly detection task/classification over images, as an image label can be simply computed as the maximum of pixel-wise anomaly predictions \cite{bergmann2021mvtec}. 
Therefore, AL is more challenging and, for most purposes, more useful compared to image-level classification, making the result explainable and actionable~\cite{tao2022deep}. In the remainder of this section, we briefly address the most relevant methods and refer readers to a survey \cite{liu_deep_2023, tao2022deep} for a broader insight into anomaly localization literature.

\myparagraph{Reconstruction-based methods.}
Most of the early machine learning methods for anomaly detection are reconstruction-based \cite{tao2022deep}, using a (variational) autoencoder~\cite{an2015variational, Bergmann2018ImprovingUD, Youkachen2019DefectSO}, or a generative adversarial network (GAN) \cite{Akay2019SkipGANomalySC, Baur2018DeepAM, Schlegl2017UnsupervisedAD} to learn to synthesize normal images. 
At inference, reconstruction errors reveal anomalies.
These methods are intuitive; however, they do not incorporate any priors on real images (\eg by pretraining), which makes them dependent on a large set of normal samples. Conversely, our method identifies anomalies with \emph{zero} normal exemplars.

\myparagraph{Deep features-based methods.}
The leading approaches in recent years belong to the class of deep feature-based methods. 
In essence, these methods leverage features extracted with the help of a larger network, pretrained on vast amounts of data, that serves as a prior. 
These embeddings have been used in various ways such as taking the $k$-nearest neighbors at image (DN2~\cite{bergman2020deep}) or sub-image level (SPADE~\cite{cohen2020sub}), creating a Teacher-Student feature-reconstruction framework~\cite{bergmann2020uninformed}, modeling the distribution of features that characterizes each pixel location as a multivariate Gaussian (PaDiM~\cite{defard2021padim}), creating a memory bank of feature patches as a representation of normality (PatchCore~\cite{roth_towards_2022}), etc. 
The intuition is that the features from the intermediate layers of a CNN trained on ImageNet~\cite{deng2009imagenet} capture higher level semantics that can be used to identify anomalies. 
As we show, our approach also benefits from using deep features.

\myparagraph{Few-shot methods.}
Few-shot anomaly detection was recently explicitly addressed with approaches such as normalizing flows \cite{rudolph2021same}, hierarchical generative models \cite{sheynin2021hierarchical}, and feature registration \cite{huang2022registration}. 
These methods, however, rely on data augmentation which is problem-specific and may require domain knowledge. 
Moreover, as observed in \cite{aota_zero-shot_2023}, they are not significantly better compared to, for example, PatchCore \cite{roth_towards_2022} which scales better with the number of samples. 
The authors of PatchCore even address the concern regarding the performance in a limited normal data setting and shows better results compared to SPADE~\cite{cohen2020sub} and PaDiM~\cite{defard2021padim}. 
Notably, we are interested in the more extreme situation where not a single \emph{normal} image is provided.

\myparagraph{Zero-shot methods.}
Zero-shot anomaly localization considers the case of anomaly detection where the anomalous regions are segmented without a set of unblemished textures to act as guidance. 
MAEDAY \cite{schwartz_maeday_2022} introduces for the first time the task of zero-shot anomaly detection. 
The method pretrains a transformer-based network which is used to reconstruct a partially masked image at inference. 
By using this in-painting network, an anomaly score can be computed by identifying the differences between the unmasked image and the reconstructed output. 
WinCLIP~\cite{Jeong_2023_CVPR} introduced a new paradigm for ZSAL using a vision-language foundation model (CLIP~\cite{radford2021learning}) which quickly gained traction~\cite{baugh2023zero, cao2023segment, chen2023zero}. These methods use text prompts to discriminate between normal and anomalous patches relying on the capacity of the multimodal foundation model to learn this distinction through large-scale training.
Aota \etal~\cite{aota_zero-shot_2023} developed a method for zero-shot anomaly detection and localization specifically for textures. 
For each pixel, the local features are averaged and compared to the $k$-nearest neighbors in the same image.
Our method is most similar to the latter as it compares local features with globally aggregated information, and it is designed to work on textures and not generic objects as \cite{schwartz_maeday_2022} and \cite{Jeong_2023_CVPR}.
Academic works that explicitly solve ZSAL only recently emerged; however, the task bears similarities to texture perception~\cite{julesz1962visual}, image saliency~\cite{cheng2014global}, texture stationarity analysis~\cite{moritz_texture_2017}, and weathering estimation~\cite{bellini2016time}.

\section{Algorithm Design}

This section describes the design decisions that went into building our method. 
We analyze how different components of a zero-shot patch-based anomaly localization system affect its performance, and we also introduce a novel procedure for estimating the anomaly degree at each spatial location.

We consider the following attributes of an AL method, identified as desirable: \emph{high sensitivity at high specificity}, \emph{ability to scale to higher resolutions}, and \emph{fast running time}. 
Importantly, we focus on a zero-shot scenario, and we are mainly interested in textures, which are largely homogeneous, save for the anomalous regions themselves. 

\begin{table*}
\centering
\begin{tabular}{l|c|c|c|c|c}
\hline
 PRO $\uparrow$ / AUROC $\uparrow$ & Colors & RandProj & Steerable & TEM & VGG \\
\hline
Moments & 46.51 / 75.62 & 40.66 / 73.33 & 64.21 / 80.78 & 53.97 / 75.64 & 61.96 / 83.82 \\
Histogram & 50.43 / 77.80 & 53.74 / 80.48 & 70.64 / 84.43 & 68.47 / 84.71 & 73.17 / 88.44 \\
SWW & 58.62 / 83.62 & 62.21 / 85.89 & \textbf{73.08} / \textbf{87.77} & 74.48 / 89.36 & 77.40 / 91.44 \\
FCA (ours) & \textbf{63.30} / \textbf{85.76} & \textbf{66.28} / \textbf{87.62} & 71.75 / 86.99 & \textbf{75.33} / \textbf{90.18} & \textbf{81.08} / \textbf{92.58} \\
\hline
\end{tabular}
\caption{Preliminary experiment, comparing our patch statistics method to different baselines. Compared in terms of two metrics: PRO(0.3) and AUROC. The best results are highlighted in bold.\belowcaptionsqueeze}
\label{tab:mvtec_128}
\end{table*}

As a generic framework for zero-shot anomaly localization, we propose the following self-similarity formulation to obtain the anomaly map $A$ from an image $I$:
%
\begin{align}
    A(x, y; F,S,R) = \sum_{\mathclap{F_r \in R(F(I))}}{S(x, y, F(I), F_r)}\;.
\end{align}
Such an AL system is defined by three different components: feature extraction ($F$), patch statistics comparison ($S$), and reference selection ($R$). 
Simply put, the anomaly score $A$ at location $(x, y)$ is computed as the sum of the costs when comparing features within one or more patches containing $(x, y)$ with a set of references $F_r$.
We note that the proposed definition is a superset of the discrete form of the stationarity measure introduced in~\cite{moritz_texture_2017}. 
While not explicitly designed for anomaly localization, by isolating the influence of each spatial location in the stationarity measure from \cite{moritz_texture_2017}, one can use it as an anomaly localization score. 
The main difference is that Moritz \etal assume the reference set $R$ consists of all patches in $F(I)$, which, as we show, is suboptimal.

\subsection{Feature Extraction}
We evaluate the effect of different feature extractors $F(I) \to \mathbb{R}^{H \times W \times C}$ and confirm the findings of previous work that pretrained neural networks provide useful features for AL.
\Cref{tab:mvtec_128} compares five feature extraction functions $F$.
The metrics used for evaluation are detailed in \Cref{sec:experiments}.
We consider using the colors directly ($F(I) = I$), convolving the image $I$ with a set of random kernels, Steerable Filters \cite{freeman1991design}, Laws' texture energy measure (TEM~\cite{Laws1980RapidTI}), and neural features from a simple pretrained VGG19 network \cite{simonyan2014very}. 
In this preliminary experiment, all feature extractors operate on a single resolution and have a small receptive field, \ie, the images are scaled to $256 \times 256$, and the feature maps have the same resolution, with $C \approx 128$ channels (except for colors, where $C = 3$). 
The random projections are inspired by \cite{elnekave2022generating}, where they are used in the context of the Sliced Wasserstein Distance, and consist of normalized random $5 \times 5$ kernels; we use only one level of steerable filters (full spatial resolution); finally, we use the concatenated output of the first two convolutional layers of the VGG network, having an effective receptive field of $5 \times 5$. 
We use a patch size of $25 \times 25$ for all stationarity measures, which is large enough to capture the difference in appearance between normal and anomalous regions.

As shown in \Cref{tab:mvtec_128}, the embeddings obtained from the VGG network consistently outperform other types of features, including traditional texture analysis methods~\cite{freeman1991design,Laws1980RapidTI}.

\subsection{Patch Statistics Comparison}

The function $S(x, y, F(I), F_r)$ evaluates the degree of anomaly at position $(x, y)$ given its local context in the feature maps $F(I)$, by comparing with the reference $F_r$. 
The function should analyze how do the local statistics around $(x, y)$ differ from the statistics in $F_r$. 
In this subsection we describe different options for $S$, together with their limitations, and introduce our Feature Correspondence Analysis (FCA) method for comparing patch statistics.

\textbf{Moments.} 
In general, only a small region around a certain location is needed to identify an anomaly. 
This leads to a trivial patch statistics comparison method, computed by averaging the features around $(x, y)$, \ie,
\begin{equation}
    S(x, y) = \biggl\|\;\frac{1}{T^2}\!\!\!\!\smashoperator[r]{\sum_{{(x', y') \in P_{xy}}}}F(I)(x',y') - \text{avg}(F_r)\;\biggr\|_2^2 \;,
\end{equation}
where $F(I)$ and $F_r$ have been omitted from $S$ for brevity, and $P_{\!xy}$ denotes a patch of size $T\times T$ centered in $(x, y)$. 
The definition can be easily extended to include spatial weighting (\eg, Gaussian) and moments of higher order, becoming equivalent to the method of moments from \cite{moritz_texture_2017} when using RGB colors directly as features.

\textbf{Histogram.} Moritz \etal~\cite{moritz_texture_2017} propose another two options for computing the stationarity measure, which can be described in our conceptual framework as using a histogram-based patch statistics comparison over RGB colors, and steerable filters, respectively. 
The histogram-based algorithm can be described as:
\begin{align}
    S(x, y) &= \text{hist}\Big(\smashoperator[r]{\bigcup_{(x', y') \in P_{\!xy}}}F(I)(x', y')\Big) \ominus \text{hist}(F_r)  \;,
\end{align}
where $\ominus$ gives the earth mover's (Wasserstein) distance between the two histograms. As in the case of moments, when computing the histogram one can employ spatial weighting to increase the importance of the pixels closer to $(x, y)$.  

\begin{figure*}
  \newcommand{\mylabel}[3]{\rlap{\hspace*{#1\linewidth}\parbox{#2}{\centering #3}\hfill\strut}}%
  \vspace*{-2ex}
  \strut\hfill\strut%
  \includegraphics[width=0.9\textwidth]{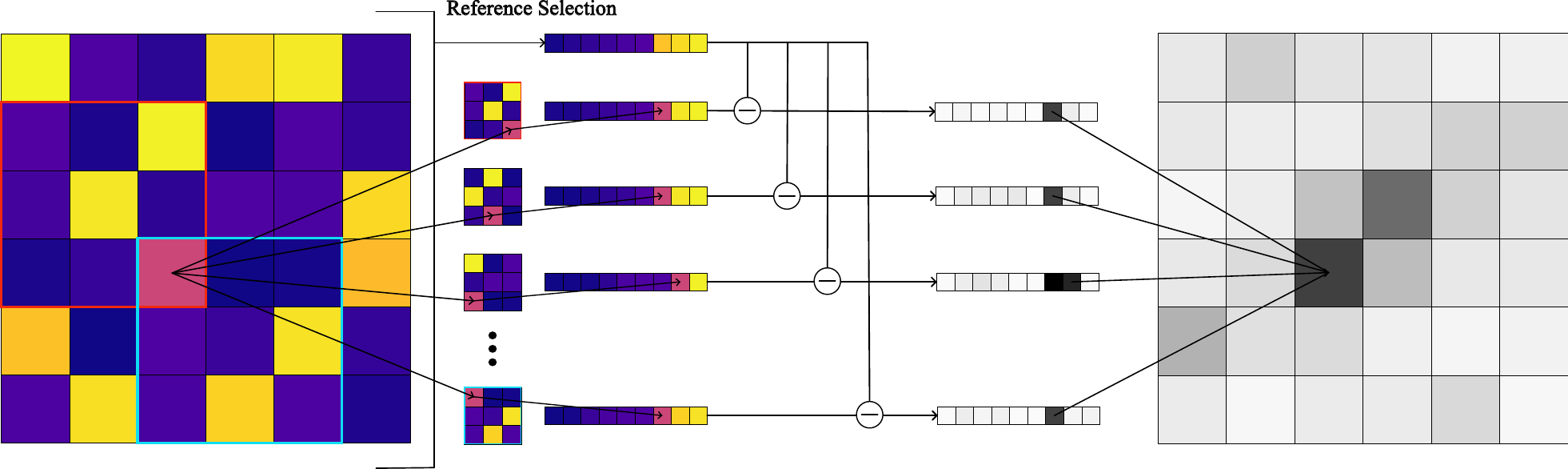}%
  \strut\hfill\strut\\[-1ex]
  {\footnotesize%
    \mylabel{0.09}{3cm}{Feature Map}%
    \mylabel{0.245}{3cm}{Patches}%
    \mylabel{0.325}{3cm}{Sorted Patch\\ Features}%
    \mylabel{0.5475}{3cm}{Non-Compliance Measures}%
    \mylabel{0.75}{3cm}{Anomaly Score $S$}
  }
  \caption{\label{fig:fca_workflow}%
    Depiction of our Feature Correspondence Analysis (FCA). All patches surrounding a pixel are compared against the \emph{reference}. The correspondence errors of the pixel in all contexts are aggregated to obtain the final anomaly score. The $\circleddash$ denotes the absolute difference.\belowcaptionsqueeze}
\end{figure*}

\textbf{Sample-weighted Wasserstein (SWW).} 
The previous methods have limited expressive powers, specifically because they consider the distribution inside a patch as a whole, unable to pinpoint ``outlieredness'' of individual samples.

That ability conveniently occurs in an efficient implementation of the 1-D Wasserstein distance 
when operating on individual samples drawn from distributions. If two sets of samples have the same size, the Wasserstein distance can be obtained by sorting the samples and then summing over the absolute differences between the elements corresponding to the same rank~\cite{elnekave2022generating}. 
That comparison of samples of the same rank within a sorting can be seen as a bijective mapping between two sample sets, and the difference between corresponding samples is an immediate measure for those samples' non-compliance with the respective other distribution.
That resulting non-compliance calculation translates directly to a coarse anomaly measure of the corresponding feature channel for each pixel; summing them across channels results in an error score $M(x,y;P)$ for each pixel $(x,y)$ in a patch~$P$.

Subsequently, we aggregate the per-patch error map $M$ into an anomaly measure for the center of the patch $P$. 
At this point, averaging $M(\cdot,\cdot;P)$ would yield the exact Wasserstein distance, and it would be equivalent to the previously defined histogram method (with bins $\to \infty$). 
Instead, we use a Gaussian-weighted average to increase the spatial sensitivity of the resulting anomaly score $S(x, y)$.

Upon cursory observation, this may resemble the weighted, sliding-window histogram calculations of Moritz \etal~\cite{moritz_texture_2017}; however, Moritz \etal compute weighted distributions before calculating their metric, whereas we preserve the original patch distribution but weight the influence of each sample's non-compliance score on the final anomaly score.
The equation for this sample-weighted Wasserstein method is:
\begin{align}
    S(x, y) &= \sum_{\mathclap{(x', y') \in P_{\!xy}}}M(x', y'; P_{\!xy})G_{\!\sigma_w}(x'-x, y'-y) \;,
\label{eq:wasserstein}
\end{align}
where $G_{\!\sigma_w}(\Delta_x, \Delta_y)$ is a spatial weighting function, for which we use a Gaussian with variance $\sigma_w^2$.
Note that we introduce this method as a conceptual bridge between comparing histograms and our FCA.

\textbf{FCA.} The previous definition of SWW allows us to separate the context size and the amount of smoothing in the aggregation through the parameter of the Gaussian;
however, the final anomaly score for any location $(x, y)$ uses as context only the patch $P_{\!xy}$.
We further leverage the bijective mapping from SWW by computing the anomaly score at location $(x, y)$ as the sum of the matching errors for position $(x, y)$ in the context of all surrounding patches, which gives:
\begin{align}
   S(x, y) &= \sum_{\mathclap{(x', y') \in P_{\!xy}}}M(x, y; P_{\!x'y'})G_{\!\sigma_p}(x'-x, y'-y)
   \;.
   \label{eq:FCA_nos}
\end{align}
Please note the change of parameters in $M$ compared to the SWW equation (\ref{eq:wasserstein}).
The main difference is that instead of considering one context patch $P_{\!xy}$ when computing $S(x, y)$, we consider all patches that contain $(x, y)$, and aggregate the contribution of the location $(x, y)$ in all of these contexts. 
Anomalies are generally considered smooth and all available datasets present anomalies as binary blobs that mark anomalous regions, rather than continuous scores depicting the contribution of each pixel to the anomaly.
To attend to this, we introduce Gaussian smoothing $\mathcal{G}_{\sigma_s}$ after matching errors, yielding the final formula:
\begin{align}
    S(x, y) = \smashoperator{\sum_{(x', y') \in P_{\!xy}}}
    \mathcal{G}_{\sigma_s} \bigl(M(\cdot,\cdot; P_{\!x'y'})\bigr)(x,y)\,
    G_{\!\sigma_p}(x'-x, y'-y)
    \,.
    \nonumber\\[-1.0\baselineskip]
   \label{eq:FCA_s}
\end{align}

The workflow of the algorithm is illustrated in~\Cref{fig:fca_workflow}. 
We name this novel method Feature Correspondence Analysis (FCA), as it computes the anomaly score based on the correspondence of features from patches to a reference.

\begin{figure}[ht]
\centering
\begin{tabular}{@{}c@{\hspace{1mm}}c@{\hspace{1mm}}c@{\hspace{1mm}}c@{}}
\includegraphics[width=0.11\textwidth]{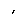} &
\includegraphics[width=0.11\textwidth]{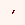} &
\includegraphics[width=0.11\textwidth]{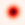} &
\includegraphics[width=0.11\textwidth]{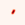}
\\
\includegraphics[width=0.11\textwidth]{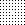} &
\includegraphics[width=0.11\textwidth]{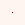} &
\includegraphics[width=0.11\textwidth]{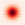} & 
\includegraphics[width=0.11\textwidth]{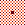}
\\
Input Image & FCA AL & Hist large $T$ & Hist small $T$
\end{tabular}
\caption{\label{fig:method_info}%
  Anomaly localization for 2 synthetic
  examples when comparing patch statistics using FCA versus histograms.
}
\end{figure}

In~\Cref{fig:method_info}, we showcase the effect of the proposed method on an artificial problem. 
We run FCA without smoothing (Equation \ref{eq:FCA_nos}) to show how our formulation allows significantly better localization of the source of the error when comparing the patch features statistics to the reference. 
While running the histogram method with a small patch size would improve the first result, it would fail in the second example because it contains a contextual (also called conditional \cite{song2007conditional}) anomaly.

\begin{figure}[ht]
\centering
\begin{tabular}{@{}c@{\hspace{1mm}}c@{\hspace{1mm}}c@{\hspace{1mm}}c@{}}
\includegraphics[width=0.11\textwidth]{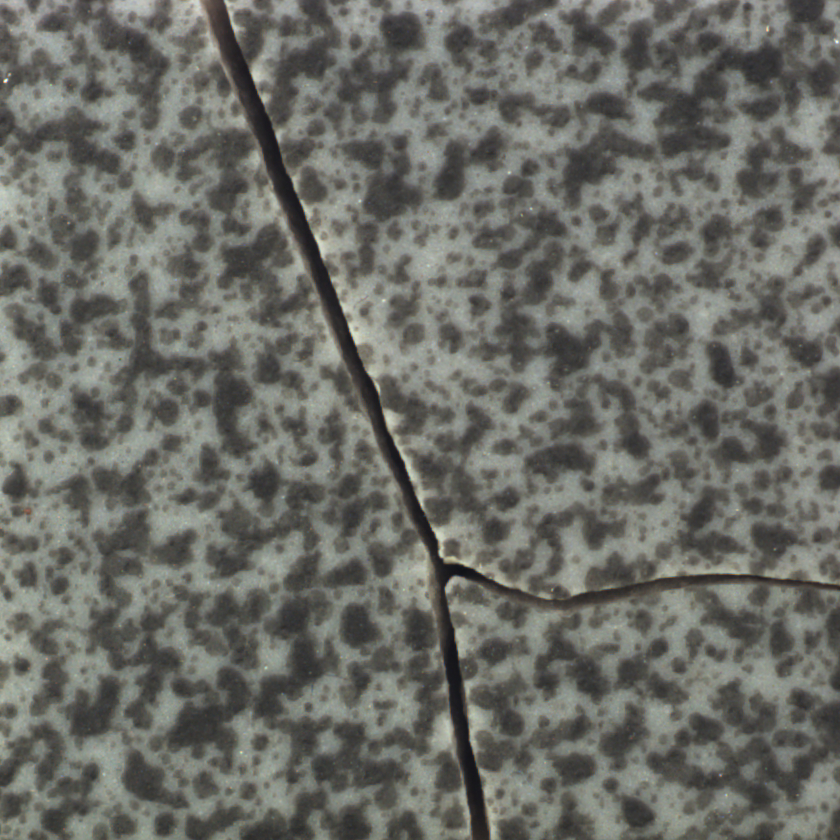} &
\includegraphics[width=0.11\textwidth]{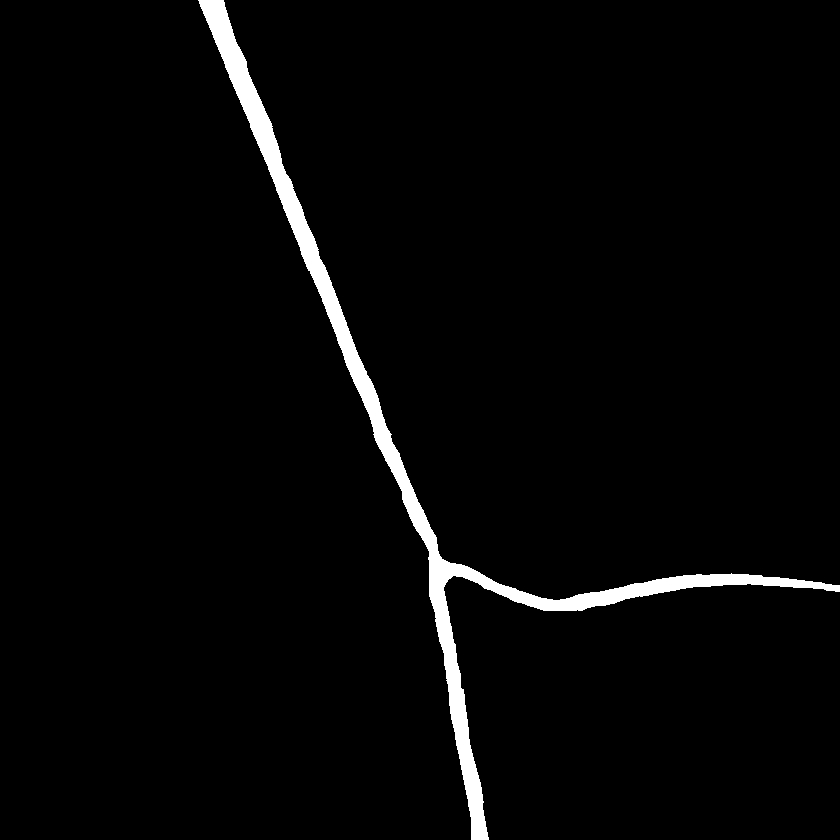} &
\includegraphics[width=0.11\textwidth]{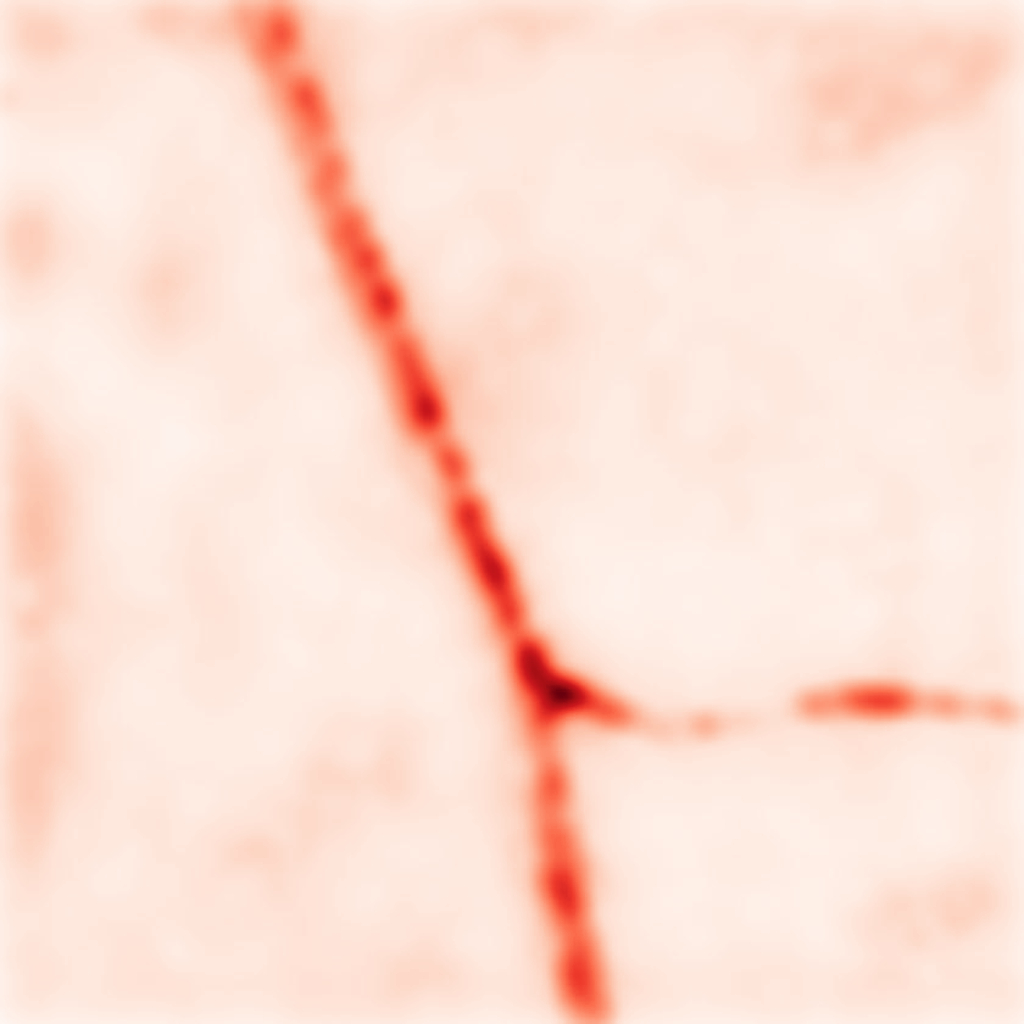} &
\includegraphics[width=0.11\textwidth]{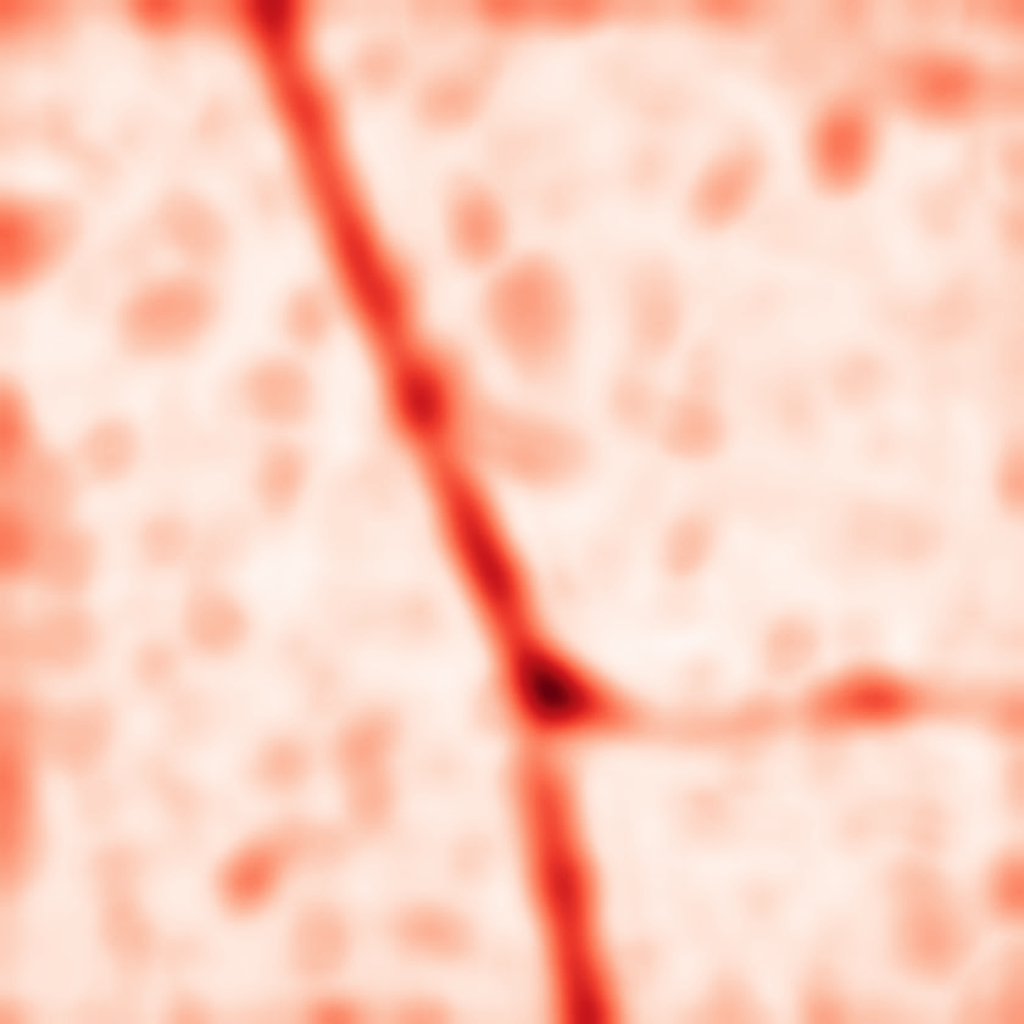} \\
\includegraphics[width=0.11\textwidth]{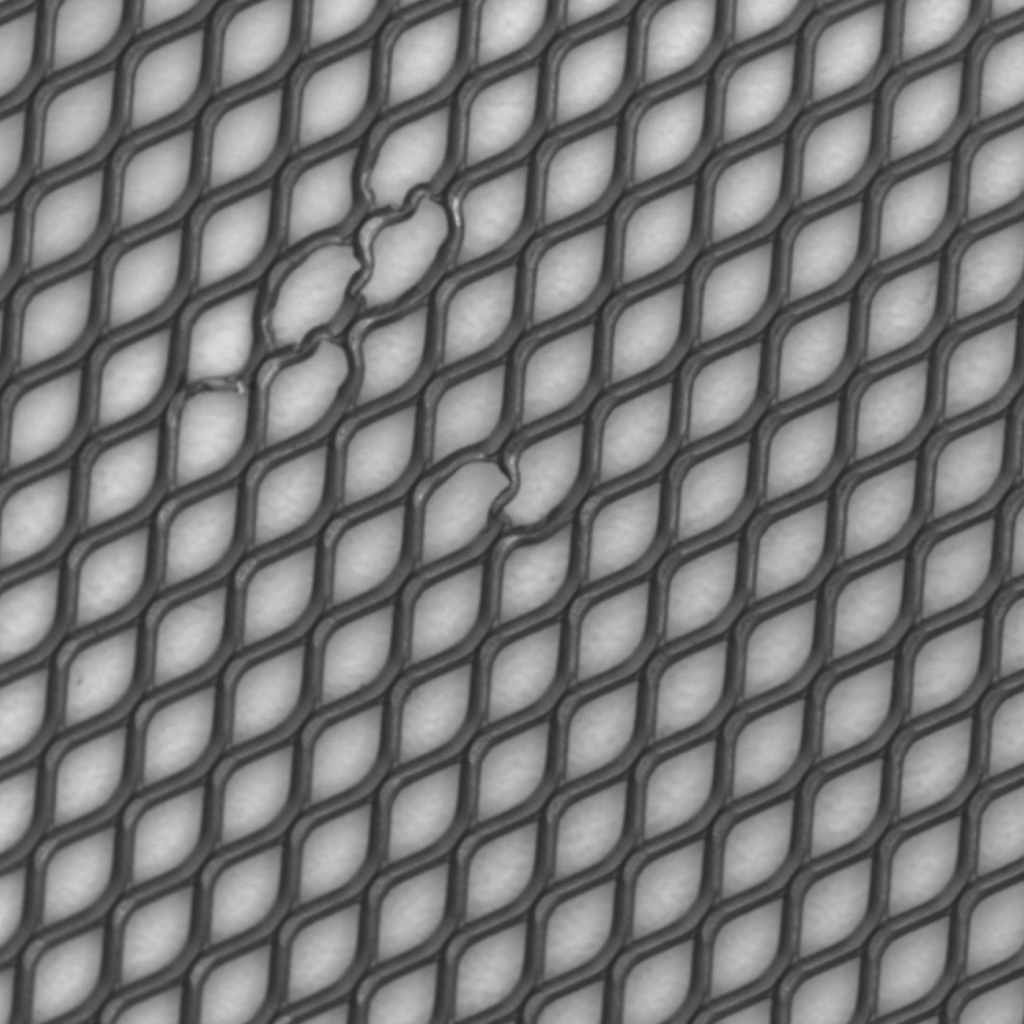} &
\includegraphics[width=0.11\textwidth]{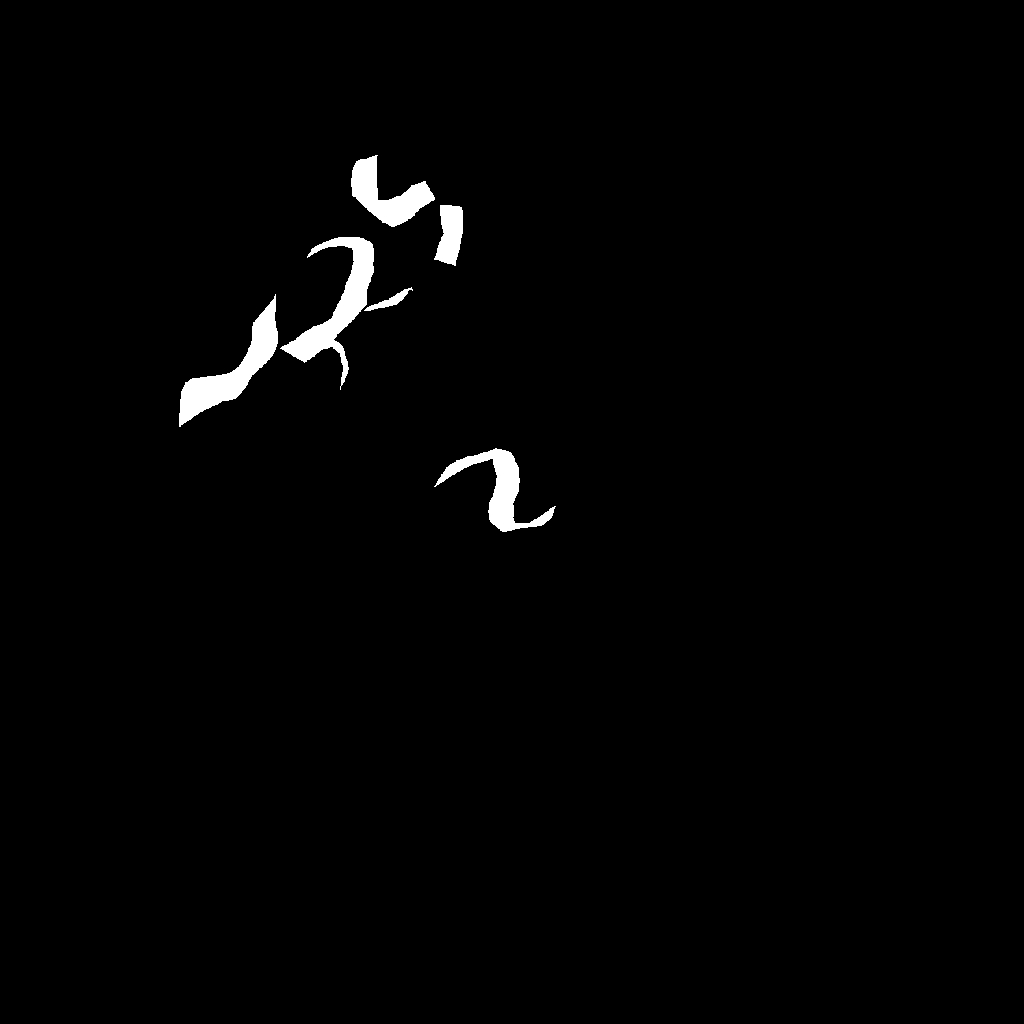} &
\includegraphics[width=0.11\textwidth]{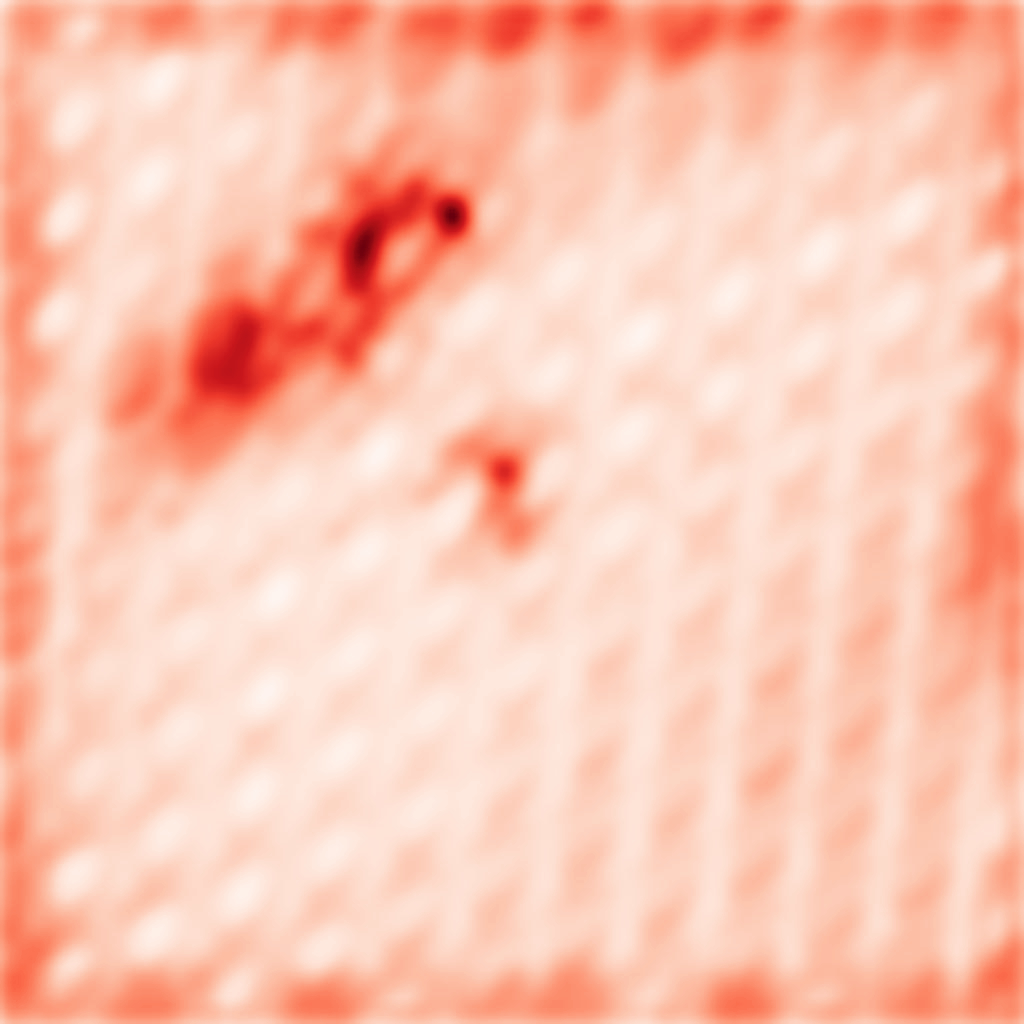} &
\includegraphics[width=0.11\textwidth]{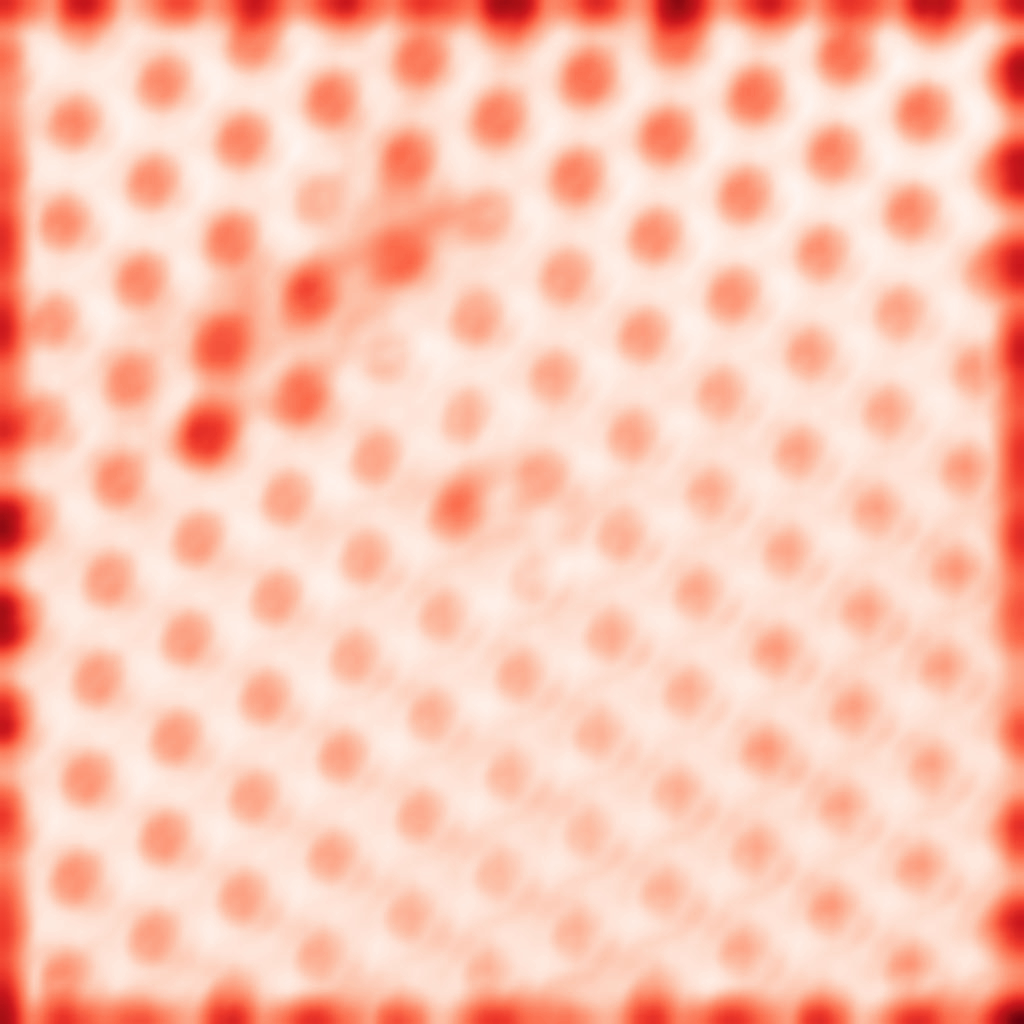} \\
Input Image & GT mask & FCA AL & Histogram AL
\end{tabular}
\caption{\label{fig:method_info_real}%
  Anomaly localization maps for 2 textures from MVTec~AD
  \cite{bergmann2019mvtec} when comparing patch statistics using
  FCA versus histograms.
  \vspace*{-0.75\baselineskip}  
}
\end{figure}

We also compare the Histogram method and our FCA on a real case example from MVTec AD \cite{bergmann2021mvtec}, in~\Cref{fig:method_info_real}.

\subsection{Reference Selection}
\label{sec:reference}

We analyze several options for the set of references $R(F(I))$. 
An intuitive solution is to use all the patches in the image as references, however, this amounts to computing the pairwise distances between all patches in an image which can be very time-consuming, scaling poorly as the image resolution increases.
Choosing a single patch at random is fast but is a poor approximation of the global statistics.

One can alternatively use one reference that aggregates the global information (\eg, global average for moments, and the histogram over the whole feature map, for the histogram-based patch statistics comparison). 
In the case of SWW and FCA, we choose $F_r$ as:
\begin{align}
    \argmin_{F_r} &\sum_{(x, y)}A(x,y; \cdot, \cdot, R=\{F_r\}) \;.
\end{align}
The feature set that minimizes the Wasserstein distance across all patches has a closed-form solution, obtained by taking the median over the features at each sorted position individually, \ie, compute the median for each order statistic for each feature channel.
We analyze the performance of the global statistic aggregation method and the trade-off between the number of random patches used and performance in~\Cref{tab:reference}.  
\begin{table}[t]
\centering
\begin{tabular}{@{}l|c|c|c@{}}
\hline
PRO / Time [s] & Hist & SWW & FCA \\
\hline
Random (1) & 60.95 / 1.1 & 59.56 / 5.7 & 62.01 / 9.2 \\
Random (3) & 67.04 / 1.1 & 66.79 / 8.2 & 69.57 / 18.7 \\
Random (10) & 72.99 / 1.2 & 73.23 / 17.3 & 75.91 / 52.3 \\
Random (100) & 74.55 / 1.7 & 75.69 / 134 & 78.45 / 482 \\
All & 74.01 / 380 & -- / 84984 & -- / 314577 \\
Global & 73.17 / 1.1 & 77.40 / 5.7 & 81.08 / 9.2 \\

\hline
\end{tabular}
\caption{\label{tab:reference}
  Analysis of the effect of the Reference Selection method. We report
  the PRO$(0.3)$ metric as well as the running time per
  image. Variants that would be unreasonably slow to be used in
  practice were marked with ``--'', and only the time was
  reported.
  \belowcaptionsqueeze
}
\end{table}

Using the median works well when the texture is homogeneous but struggles to capture the global statistics for multimodal textures (\eg, structured textures with the period larger than the patch size). 
To avoid this issue, one can use the pairwise distances and discard the outliers by considering only the closest $k$ distances.
In this case, $R$ selects the $k$-nearest neighbors (\knn) over all patches in the feature maps, with respect to the cost $S(x,y; F(I), F_r)$. 
In~\Cref{sec:sub_results}, we only report results using \knn references when running on low-resolution feature maps, due to the high running time of this method.
We note that employing a WideResnet-50~\cite{BMVC2016_87} as feature extractor, using the first moment for patch statistics comparison, and taking the \knn for reference selection yields a system equivalent to Aota \etal~\cite{aota_zero-shot_2023}.

\subsection{Final Method and Implementation Details}

In accordance with the observations made in this section, we design our final anomaly localization system to use neural features from a pretrained neural network, evaluate the local statistics with the newly introduced FCA, and use the median for reference selection as a balance between fidelity and speed. 
Following recent work on anomaly detection \cite{cohen2020sub, defard2021padim, roth_towards_2022}, we use a WideResnet-50 network \cite{BMVC2016_87} and extract the features computed by the second convolutional block, yielding feature maps with $512$ channels. 
Because the output of this block has a resolution 8 times smaller than the input, and FCA can handle relatively large context sizes, we choose to run the method at full resolution and not resize it as a preprocessing step as done in previous work \cite{aota_zero-shot_2023, defard2021padim, roth_towards_2022}.  
All patch statistics comparison variants, including our FCA, have been implemented in PyTorch \cite{paszke2019pytorch}, utilizing CUDA acceleration, and ran on an NVIDIA RTX A5000 GPU.
We use the same hyperparameters for all experiments, setting $\sigma_p = 3.0$, $\sigma_s = 1.0$. 
The patch size $T$ should be set depending on the size of the feature maps. 
We use $T = 9$ when running at full dataset resolution and $T = 3$ for consistency with Aota \etal~\cite{aota_zero-shot_2023} when running at $320 \times 320$.

\section{Experiments}
\label{sec:experiments}

We compare our approach with state-of-the-art methods in zero-shot anomaly detection as well as a few other adapted baselines. 
Several datasets are considered in order to assess the robustness of the proposed approach.

\subsection{Datasets}
\label{sec:datasets}

\myparagraph{MVTec AD.} Currently, the dataset most used in the context of anomaly detection is the MVTec~AD dataset~\cite{bergmann2021mvtec, bergmann2019mvtec}. 
We use the 5 texture classes, accumulating over 500 test images and their (manually annotated) segmentation masks. 
The resolution of these images ranges from $840 \times 840$ to $1024 \times 1024$ pixels. 
Past works \cite{aota_zero-shot_2023, bergmann2021mvtec, roth_towards_2022} propose various preprocessing and postprocessing setups,
consisting of resizing and cropping to various resolutions. 
For a fair evaluation, we compute the metrics at full resolution, following the original evaluation script from the dataset provider \cite{bergmann2021mvtec}. 
The only adaption performed is cropping to the center before evaluation to avoid computing metrics on the edges of the images where most methods do not provide reliable scores \cite{aota_zero-shot_2023}.

\myparagraph{Woven Fabric Textures.} Bergmann \etal \cite{Bergmann2018ImprovingUD} introduced a small dataset for the task of defect segmentation containing two woven fabric textures (denoted WFT from here on). For each of them, 50 test images and segmentation masks are provided. 
The resolution of the images is $512 \times 512$.

\myparagraph{DTD-Synthetic.} Aota \etal \cite{aota_zero-shot_2023} constructed an artificial dataset to evaluate anomaly detection methods on more diverse data, including anisotropic textures. The dataset is based on the Describable Texture Dataset \cite{cimpoi2014describing} on which various types of defects were artificially added. The textures are also randomly rotated and cropped, eventually yielding $1304$ images of small resolution ($180 \times 180$ to $384 \times 384$). Following \cite{aota_zero-shot_2023}, all images are resized to a fixed $320 \times 320$.

\myparagraph{Aitex.} The Aitex dataset \cite{silvestre2019public} contains uniform fabric textures. The defects have been manually annotated in the original images of size $256 \times 4096$. Following standard practice, we split the images into square pieces. 
Additionally, we discard all frames that are not completely covered by the texture and images that do not contain any anomaly.
For consistency with \cite{aota_zero-shot_2023}, we resize the images to $320 \times 320$.

\subsection{Metrics}

The main metric for anomaly localization is the threshold-independent AUROC (area under the receiver operating characteristic curve). 
This metric is not very sensitive to spatially small anomalies. 
To account for this, \cite{bergmann2019mvtec} introduced the PRO(0.3) metric which weighs the size of each anomalous region, and only computes the integral up to a False Positive Rate of $0.3$. 
Since the purpose of the proposed method is to obtain a more detailed anomaly segmentation, the PRO metric is our most important indicator.
We additionally report the pixel-level maximum $F_1$ score~\cite{chinchor1992muc, zou2022spot}, corresponding to the ($F_1$-)optimal threshold.
Our contribution deals with anomaly localization and does not focus on image-level labels (computed as the maximum across the anomaly scores map). 
Therefore, the anomaly classification metric, AUROC$_c$, is only reported on the MVTec AD dataset and omitted for other experiments.

\subsection{Results}
\label{sec:sub_results}

We compare our final method against several existing methods for zero-shot and few-show anomaly localization.
The results on the MVTec AD dataset are reported in~\Cref{tab:mvtec_all} below. 
We compare our system against MAEDAY, an image-reconstruction-based zero-shot method~\cite{schwartz_maeday_2022}, WinCLIP~\cite{Jeong_2023_CVPR}, SAA~\cite{cao2023segment}, and April-GAN~\cite{chen2023zero} based on visual-language models, and Aota \etal~\cite{aota_zero-shot_2023} which employs a WideResnet as feature extractor, and uses a simple average for patch statistics comparison, combined with a \knn search. Despite being multi-shot methods, we include PatchCore~\cite{roth_towards_2022} and RD++~\cite{Tien_2023_CVPR} for reference.
We additionally adapt methods that were not explicitly designed for ZSAL but are related in scope, for a more complete comparison. 
Bellini \etal \cite{bellini2016time} propose a method for weathering arbitrary textures from a single image, and uses an age-estimation procedure as the first step in their pipeline. 
The age-estimation procedure targets the same goal, to highlight regions in an image that stray away from the pristine appearance. 
Saliency-RC~\cite{cheng2014global} as a saliency detection method highlights parts of the image that stand out, which is related to anomaly localization. However, as mentioned by the authors, the method's performance on textures is limited.

\begin{table}[ht]
\centering
\begin{tabular}{l|c|c|c}
\hline
 & PRO (0.3) & AUROC & AUROC$_c$ \\
\hline

\color{Gray} PatchCore all$^\dag$ & \color{Gray} 93.64 & \color{Gray} 97.52 & \color{Gray} 98.96 \\

\color{Gray} RD++ all$^\dag$~\cite{Tien_2023_CVPR} & \color{Gray} 96.06 & \color{Gray} 98.06 & \color{Gray} 99.80 \\
\hline
Saliency \cite{cheng2014global} & 22.92 & 58.41 & 46.92 \\
Bellini \etal \cite{bellini2016time} & 50.75 & 76.06 & 36.90 \\
MAEDAY$^\dag$ \cite{schwartz_maeday_2022} & -- & 75.20 & 88.90 \\
WinCLIP$^\dag$~\cite{Jeong_2023_CVPR} & 71.5 & 89.06 & 99.64 \\
SAA+~\cite{cao2023segment} & 64.79 & 77.82 & 93.86 \\
April-GAN~\cite{chen2023zero} & 92.57 & 96.51 & 97.61 \\ 
Aota \etal \cite{aota_zero-shot_2023} & 93.82 & 97.47 & \textbf{99.67} \\
\hline
Ours$_{320}$ & 95.46 & 97.74 & 99.21 \\
Ours$_{320}$ + \knn & 95.58 & 97.77 & 99.17 \\
Ours & \textbf{97.18} & \textbf{98.73} & 99.58 \\
\hline
\end{tabular}
\caption{\label{tab:mvtec_all}%
  Quantitative comparison on MVTec~AD. 
  We note with~$^\dag$ results taken from different papers (may be evaluated slightly differently, as discussed in~\Cref{sec:datasets}). 
  The subscript $_{320}$ marks running our method at the lower
  resolution. 
  PatchCore and RD++ are included for reference despite being multi-shot methods.
  \belowcaptionsqueeze
  }
\end{table}
%


\begin{figure*}[ht]
\setlength{\tabcolsep}{1pt}
\renewcommand{\arraystretch}{0.7}
\newlength{\imw}
\setlength{\imw}{0.15\textwidth}
\centering
\begin{tabular}{@{}c@{\hspace{2mm}}c@{\hspace{1mm}}c@{\hspace{1mm}}c@{\hspace{1mm}}c@{\hspace{1mm}}c@{\hspace{1mm}}c@{}}

\multirow{2}{*}[3em]{\rotatebox[origin=c]{90}{ MVTec AD}}
 & \includegraphics[width=\imw]{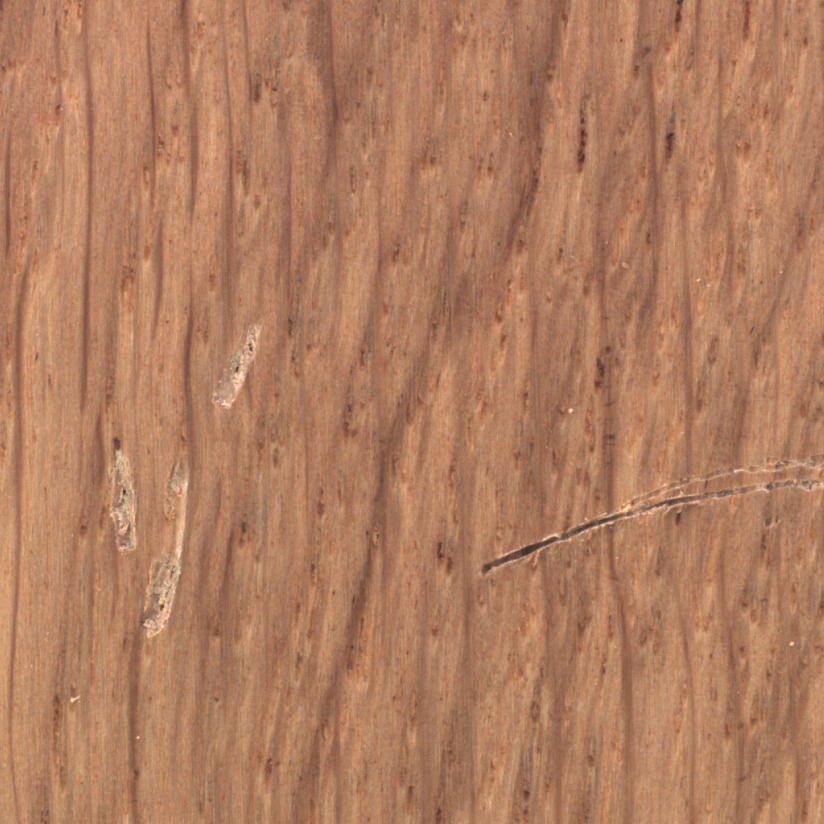} &
\includegraphics[width=\imw]{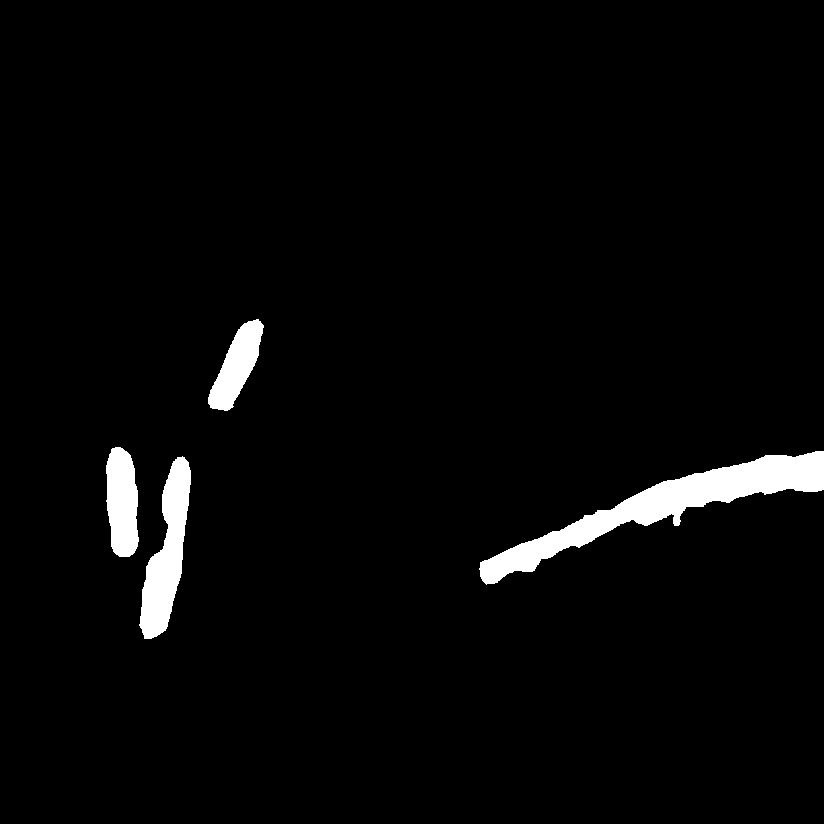}  &
\includegraphics[width=\imw]{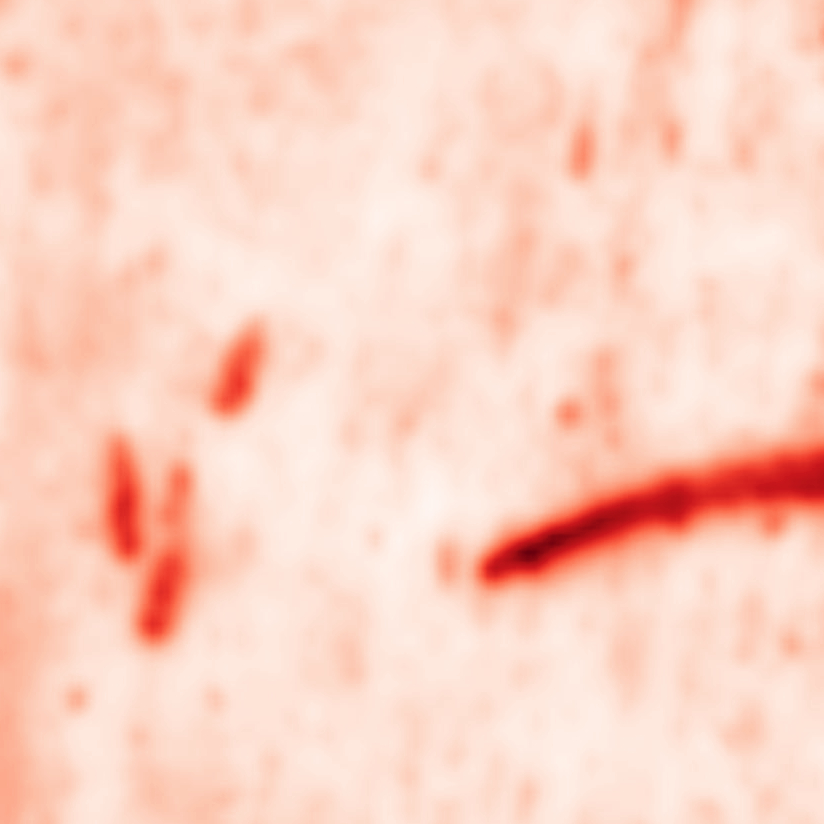}  &
\includegraphics[width=\imw]{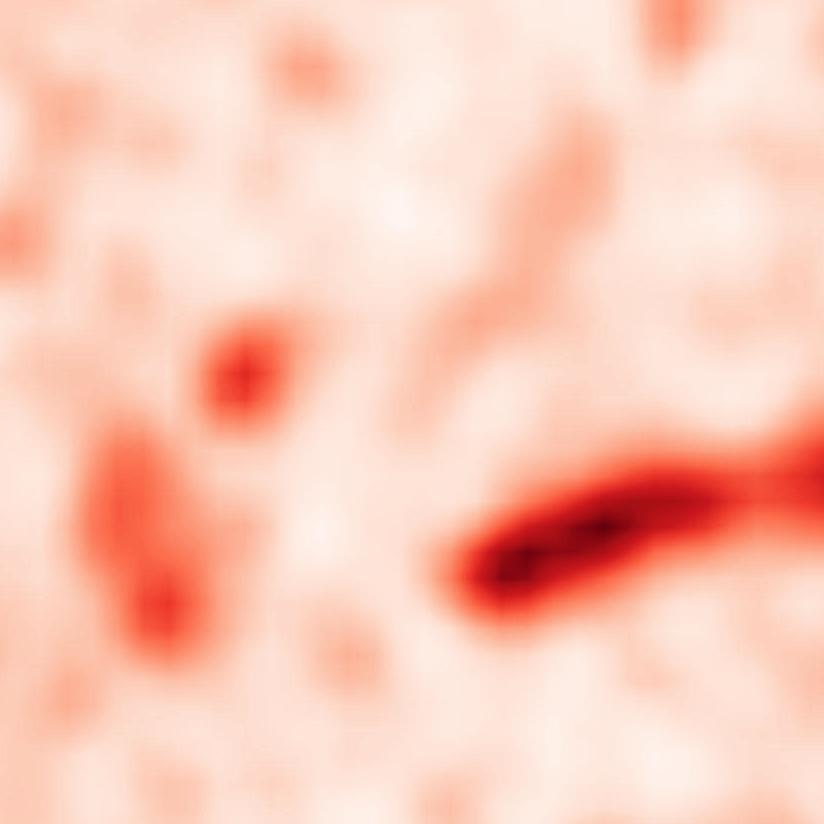}  &
\includegraphics[width=\imw]{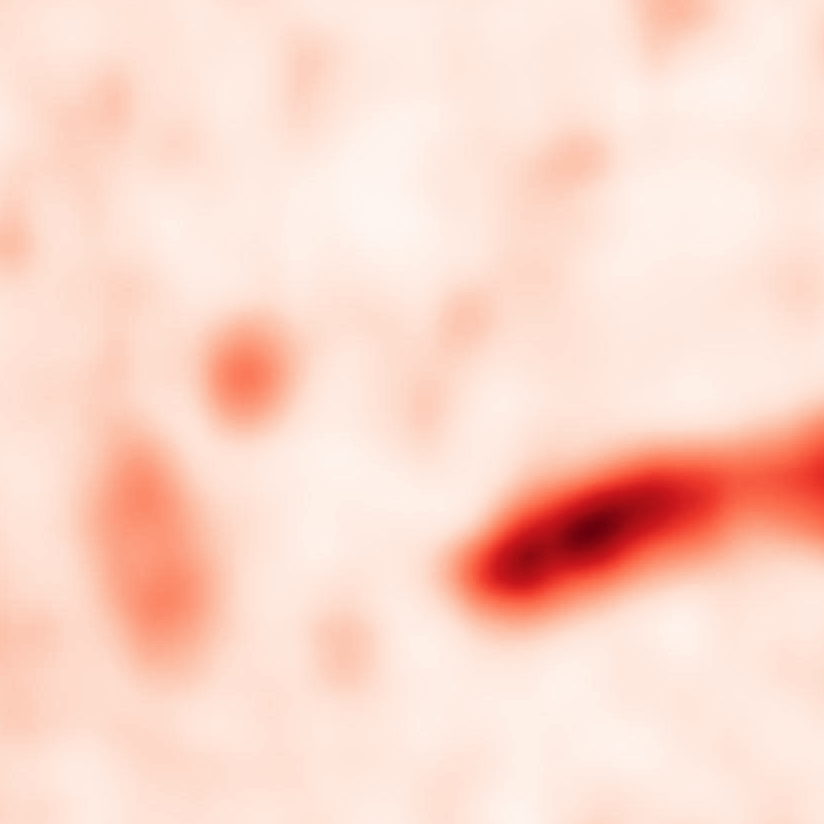}  &
\includegraphics[width=\imw]{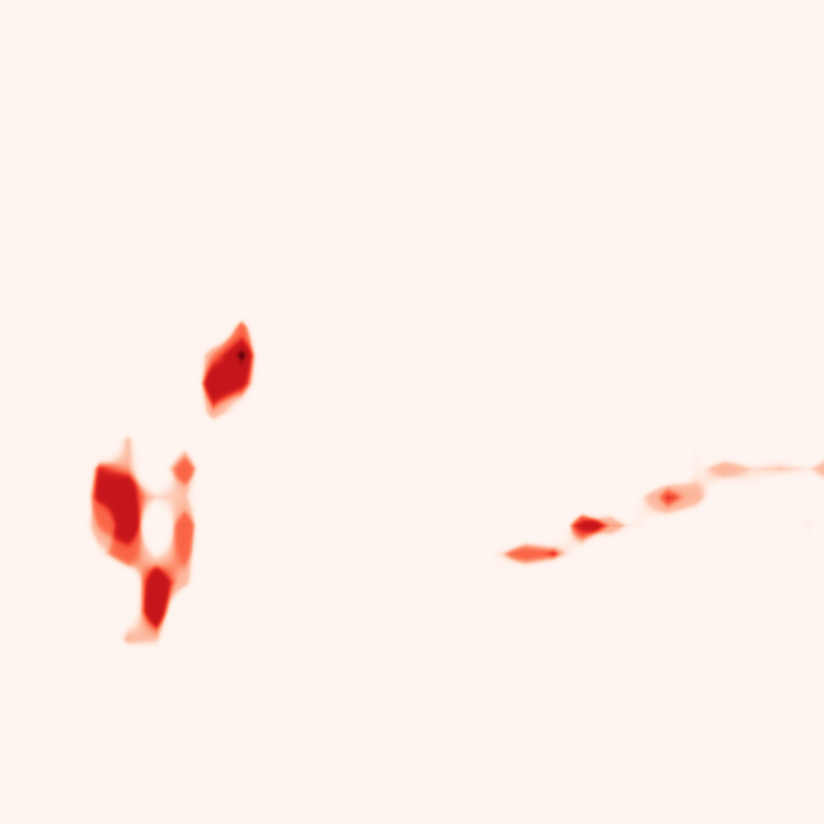} \\

& \includegraphics[width=\imw]{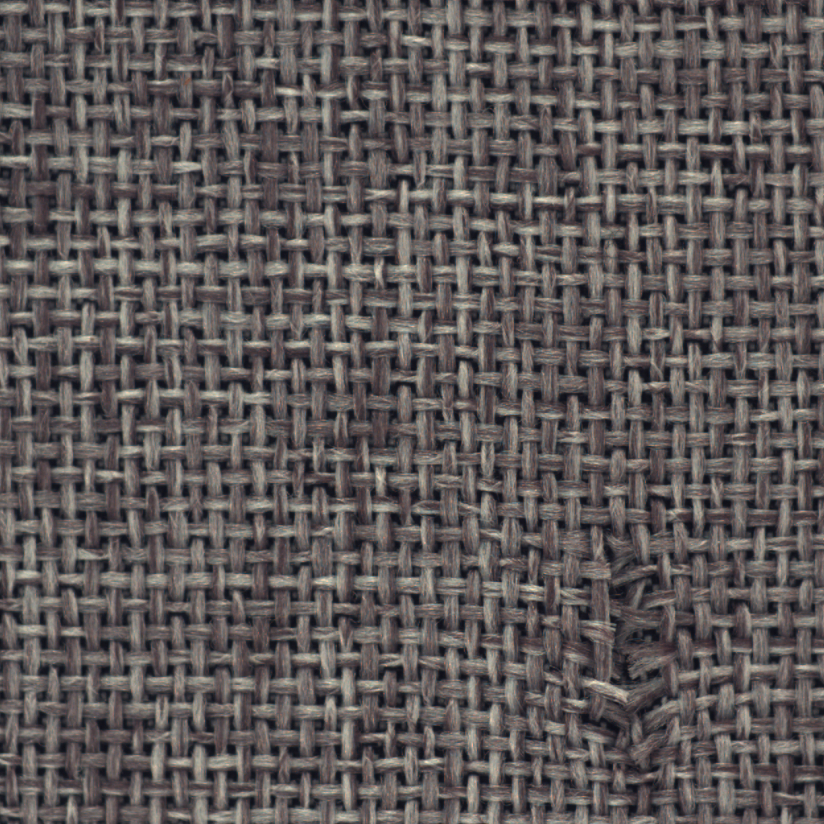} &
\includegraphics[width=\imw]{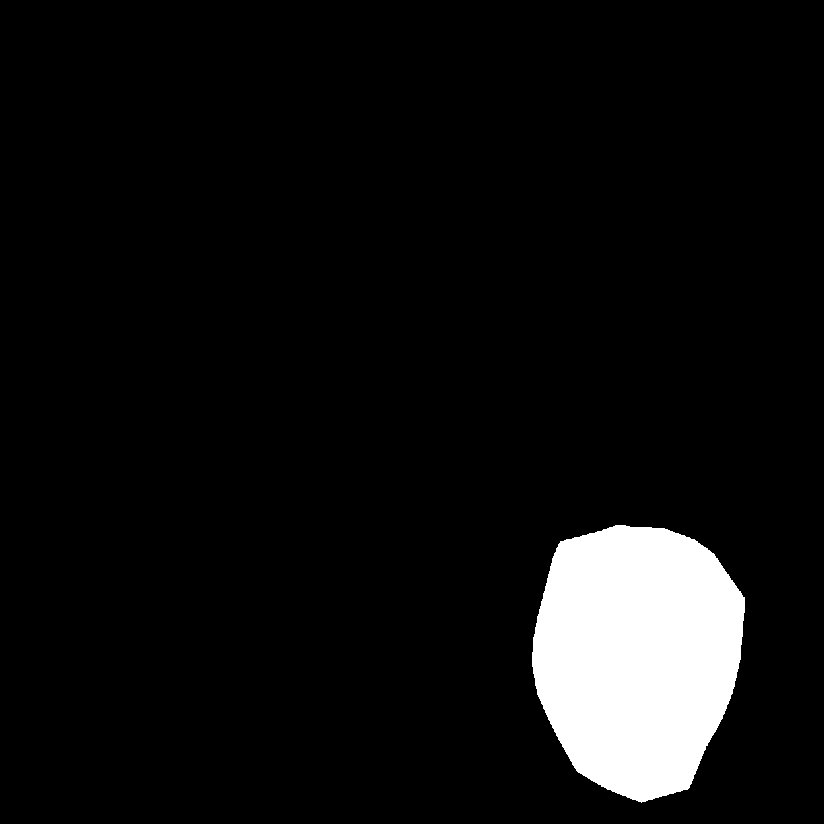}  &
\includegraphics[width=\imw]{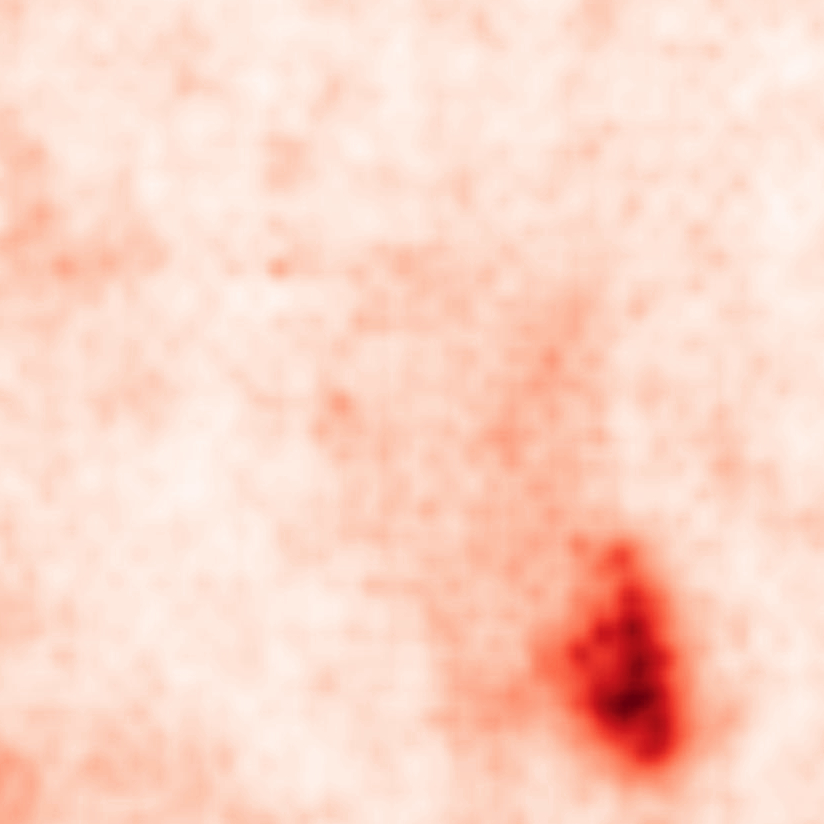}  &
\includegraphics[width=\imw]{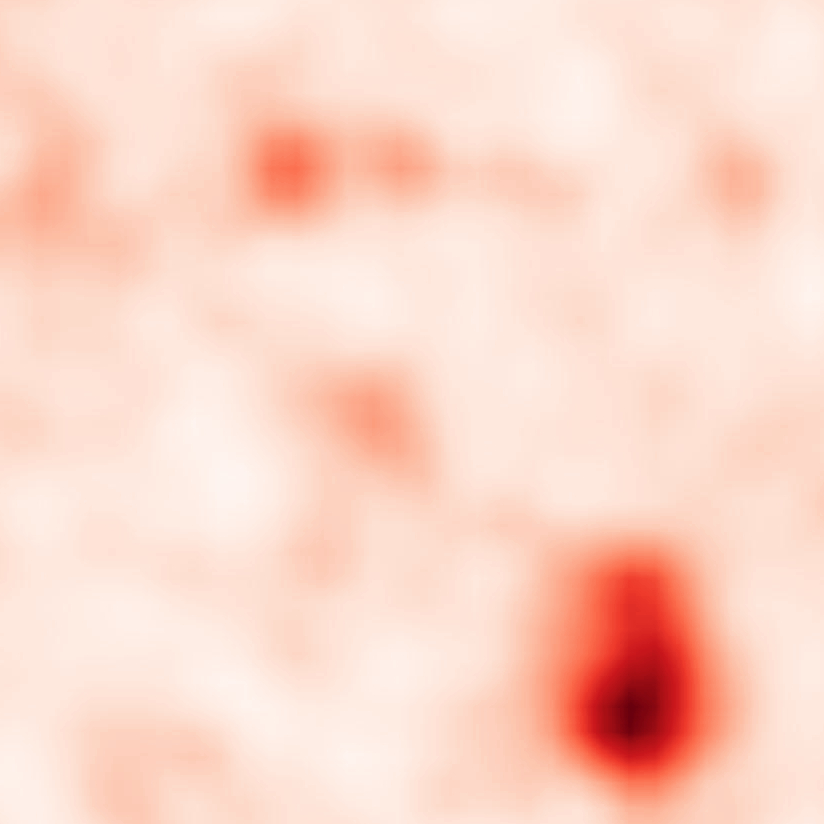}  &
\includegraphics[width=\imw]{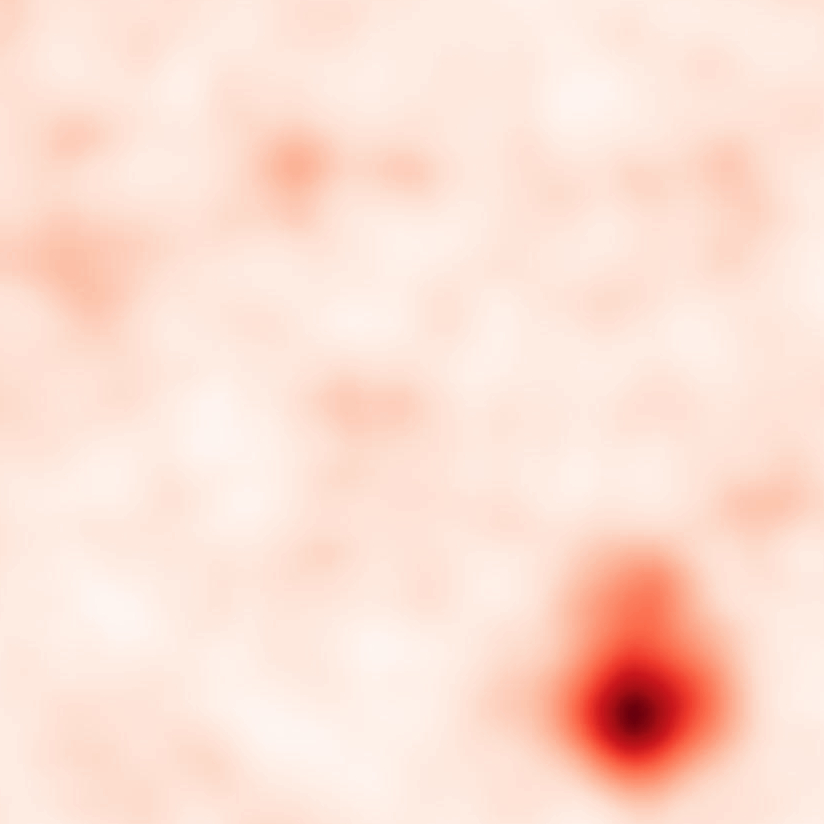}  &
\includegraphics[width=\imw]{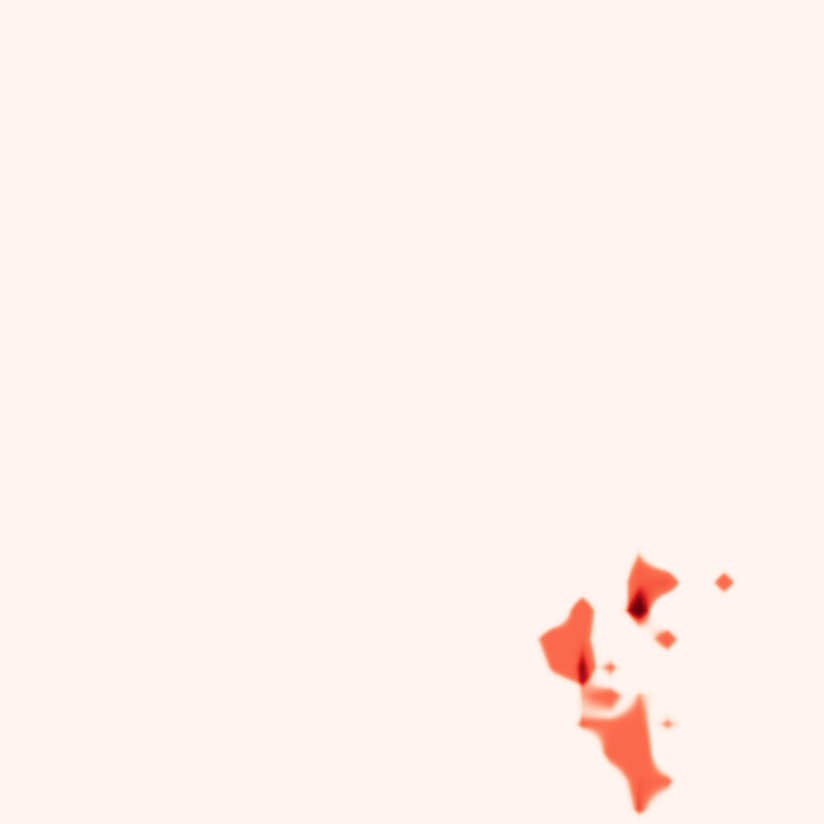}  \\

\multirow{2}{*}[2em]{\rotatebox[origin=c]{90}{DTD-synthetic}}
& \includegraphics[width=\imw]{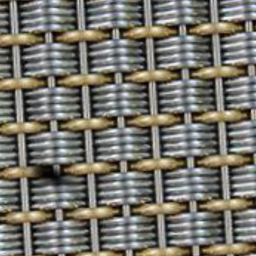} &
\includegraphics[width=\imw]{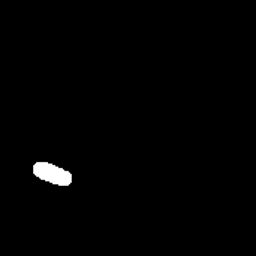}  &
\includegraphics[width=\imw]{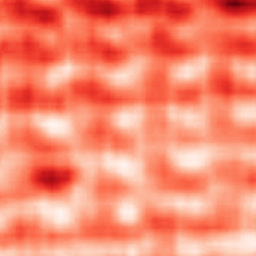}  &
\includegraphics[width=\imw]{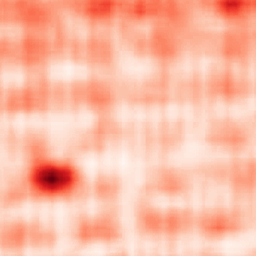}  &
\includegraphics[width=\imw]{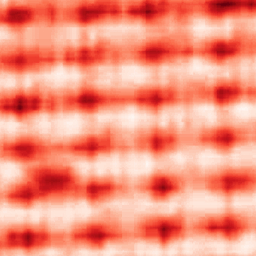}  &
\includegraphics[width=\imw]{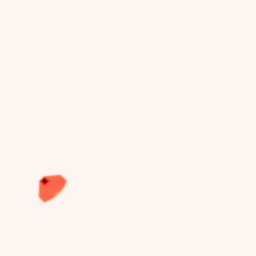}  \\

& \includegraphics[width=\imw]{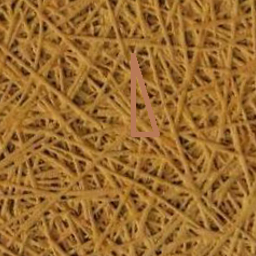} &
\includegraphics[width=\imw]{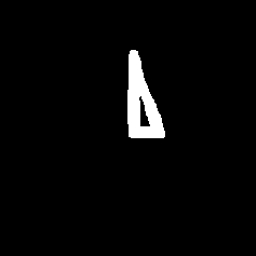}  &
\includegraphics[width=\imw]{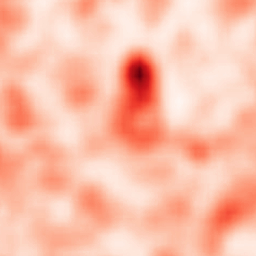}  &
\includegraphics[width=\imw]{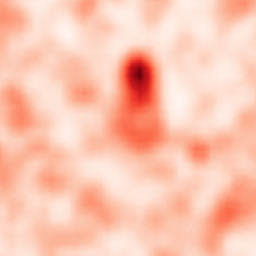}  &
\includegraphics[width=\imw]{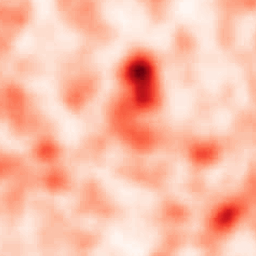}  &
\includegraphics[width=\imw]{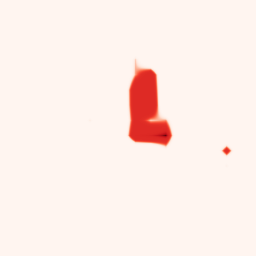}  \\

\multirow{2}{*}[3em]{\rotatebox[origin=c]{90}{Woven Fabric Textures}}
& \includegraphics[width=\imw]{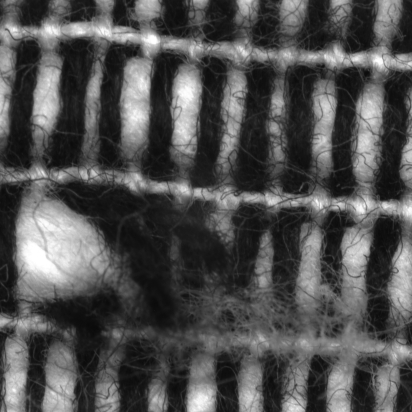} &
\includegraphics[width=\imw]{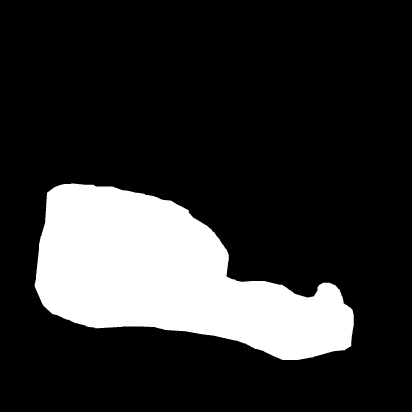}  &
\includegraphics[width=\imw]{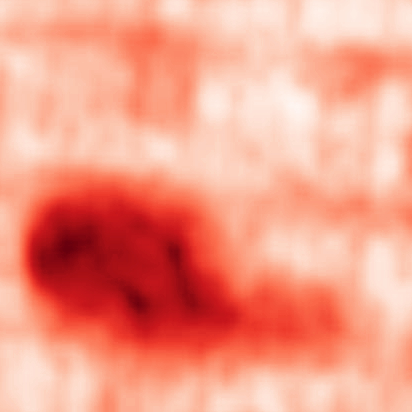}  &
\includegraphics[width=\imw]{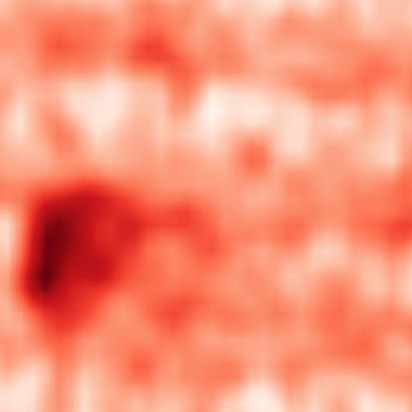}  &
\includegraphics[width=\imw]{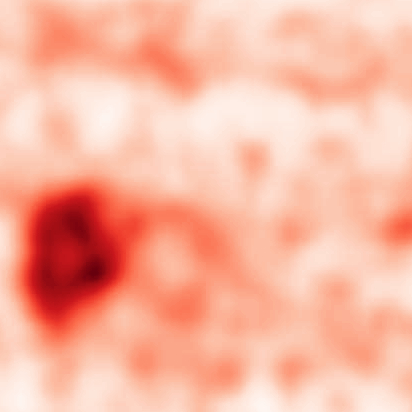}  &
\includegraphics[width=\imw]{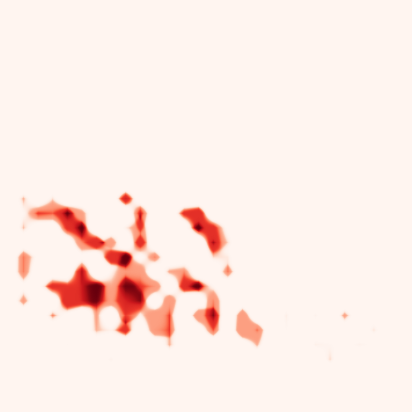}  \\

& \includegraphics[width=\imw]{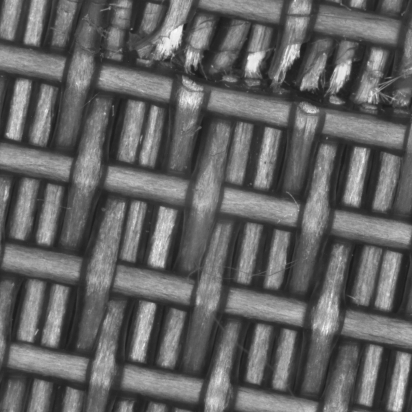} &
\includegraphics[width=\imw]{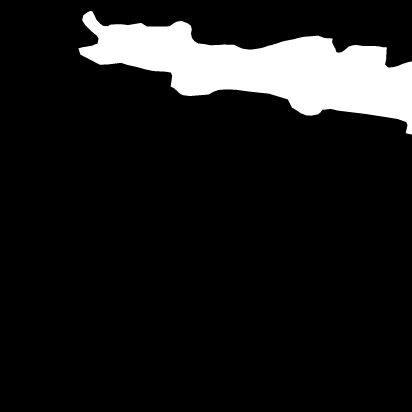}  &
\includegraphics[width=\imw]{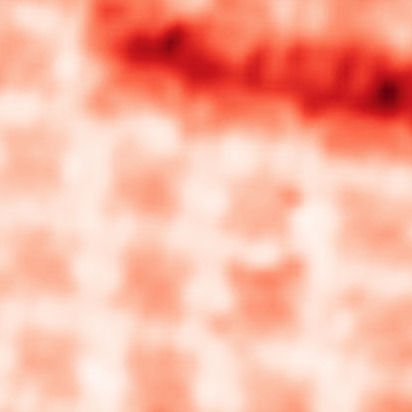}  &
\includegraphics[width=\imw]{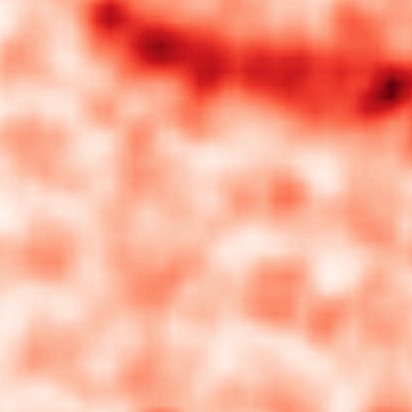}  &
\includegraphics[width=\imw]{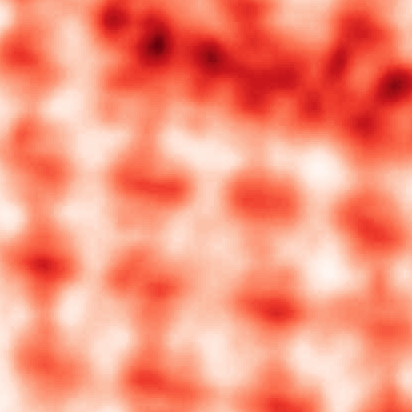}  &
\includegraphics[width=\imw]{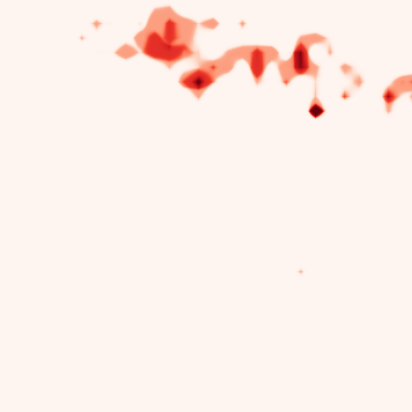}  \\

\multirow{1}{*}[5em]{\rotatebox[origin=c]{90}{Aitex}}
& \includegraphics[width=\imw]{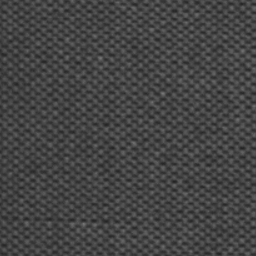} &
\includegraphics[width=\imw]{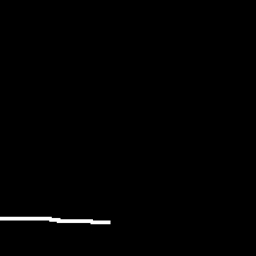}  &
\includegraphics[width=\imw]{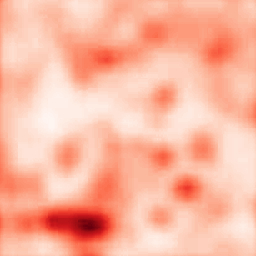}  &
\includegraphics[width=\imw]{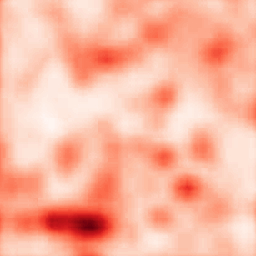}  &
\includegraphics[width=\imw]{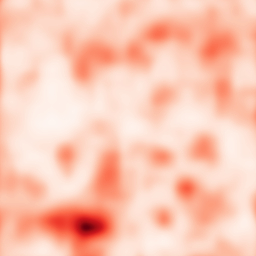}  &
\includegraphics[width=\imw]{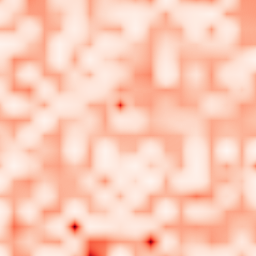}  \\

& Input Image & GT mask & Ours & Ours$_{320}$ + \knn & Aota \etal \cite{aota_zero-shot_2023} & April-GAN~\cite{chen2023zero}
\end{tabular}
\caption{Qualitative comparison on challenging examples. 
The images are shown after cropping to center.
}
\label{fig:qualitative}
\end{figure*}


The proposed method improves the localization of anomalous textures significantly compared to the previous state-of-the-art zero-shot method of Aota \etal by using a more precise method for comparing patch statistics. 
Moreover, our system even outperforms prominent multi-shot anomaly detection methods PatchCore and RD++.
April-GAN localizes the anomalies significantly better than methods following the same paradigm (WinCLIP, SAA+), which is consistent with April-GAN having won the zero-shot visual anomaly and novelty detection challenge at CVPR 2023. 
Nonetheless, on textures the method is clearly outperformed by our approach.

\begin{table}[ht]
\centering\vspace*{-0.5\baselineskip}
\begin{tabular}{l|c|c|c}
& PRO (0.3) & AUROC & F$_1$ \\
\hline
\multicolumn{4}{l}{DTD-Synthetic }\\
\hline

April-GAN~\cite{chen2023zero} & 88.50 & 95.32 & 52.40 \\ 
Aota \etal \cite{aota_zero-shot_2023} & 94.32 & 98.00 & 65.96 \\
Ours & 94.71 & 98.03 & 69.87 \\
Ours + \knn & \textbf{95.93} & \textbf{98.51} & \textbf{71.79} \\
\hline
\multicolumn{4}{l}{WFT}\\
\hline
April-GAN~\cite{chen2023zero} & 84.97 & 94.90 & 71.51 \\ 
Aota \etal \cite{aota_zero-shot_2023} & 84.59 & 96.11 & 72.07 \\
Ours$_{320}$ & 73.54 & 93.09 & 65.86 \\
Ours$_{320}$ + \knn & 86.24 & 96.19 & 72.76 \\
Ours & \textbf{89.57} & \textbf{98.26} & \textbf{79.13} \\
\hline
\multicolumn{4}{l}{Aitex}\\
\hline
April-GAN~\cite{chen2023zero} & 72.62 & 85.90 & 35.23 \\ 
Aota \etal \cite{aota_zero-shot_2023} & 87.11 & 96.70 & 61.09 \\
Ours & 91.07 & 97.51 & 62.39 \\
Ours + \knn & \textbf{91.24} & \textbf{97.52} & \textbf{62.62} \\
\end{tabular}
\caption{\label{tab:rest_3_quant}
  Quantitative comparison on DTD-Synthetic, Woven Fabric Textures
  (WFT), and Aitex.
  \belowcaptionsqueeze
}
\end{table}

We additionally run experiments on the DTD-Synthetic dataset \cite{aota_zero-shot_2023}, Woven Fabric Textures \cite{Bergmann2018ImprovingUD}, and Aitex \cite{silvestre2019public}. 
In these experiments we only compare with the identified leading methods in ZSAL, \ie, \cite{aota_zero-shot_2023} and \cite{chen2023zero}. The results are presented in~\Cref{tab:rest_3_quant} and show that our method consistently improves upon prior art in all metrics. 
We find that using \knn for reference selection can improve our results at the cost of a higher running time. 
Importantly, as the resolution increases, the value added by our FCA also grows. 
This can be seen in the results on the WFT ($512 \times 512$) and MVTec AD ($1024 \times 1024$) datasets where running our method at full resolution outperforms the lower resolution + \knn variant.

In~\Cref{fig:qualitative} we present a qualitative comparison to the leading zero-shots methods~\cite{aota_zero-shot_2023,chen2023zero}. 
We show the anomaly predictions on challenging samples from each dataset. 
Compared to \cite{aota_zero-shot_2023}, the anomaly maps produced by our method have higher fidelity, with more precise localization (rows 1, 2, 4), fewer false positives (rows 3, 6), and more complete coverage of the anomalous regions (rows 5, 7). 
April-GAN generally fails to detect the entire anomaly.
For more visualizations please see the supplementary material.

In addition to the study of various design choices in our system with respect to feature extraction and patch statistics comparison (\Cref{tab:mvtec_128}), reference selection (\Cref{tab:reference}), image size and \knn usage (\Cref{tab:mvtec_all,tab:rest_3_quant}), we present a direct ablation and a sensitivity analysis for our parameters $T$, $\sigma_p$ and $\sigma_s$ in the supplementary material. We also include there a brief analysis of possible failure cases for our method.

\section{Discussion}

The results suggest the proposed method predicts anomaly scores with high fidelity.
While FCA generally performs better compared to other methods, it has a relatively high complexity.
Computing local moments or histograms can be done efficiently thanks to the separability of the Gaussian kernel.
This does not apply to FCA which requires sorting the values inside each sliding window.
\Cref{tab:complexity} reports the computational complexity and running time of patch-comparison-based methods for anomaly localization.
The summary shows that despite the added complexity, our method scales sensibly with image resolution.

\begin{table}[ht]
\centering\vspace*{-0.5\baselineskip}
\setlength{\tabcolsep}{5pt}
  \begin{tabular}{l|c|c}
    \hline
    Method & Complexity & Time [s]\\
           &            & {\footnotesize $\;320\,\times\,320$}\;/\,{\footnotesize $1024\!\times\!1024$}\\
    \hline
    Moments & $O(NTD)$ & 0.04 / 0.05 \\
    Histogram & $O(NTBD)$ & 0.06 / 0.08 \\
    Aota \etal & $O(NTD + N^2D)$ & 1.10 / 199$^\ddagger$ \\
    Ours & $O(NT^2\log(T)D)$ & 0.71 / 1.07 \\
    Ours + \knn & $O(N^2T^2\log(T)D)$ & 3.97 / 392$^\ddagger$ \\
    \hline
  \end{tabular}
  \caption{\label{tab:complexity}%
    Complexity analysis. $D$: number of features; $T^2$: patch area; $N$: image pixel count; $B$: bins per histogram. 
    Inference time is computed at $320\times320$ and $1024\times1024$ resolutions.\\
    $^\ddagger$Aota~\etal and Ours + \knn scale poorly with large image sizes.
    \vspace*{-1ex}
    \belowcaptionsqueeze}
\end{table}

\begin{figure}[H]
\centering\vspace*{-0.5ex}
\begin{tabular}{@{}c@{\hspace{1mm}}c@{\hspace{1mm}}c@{\hspace{1mm}}c@{}}
\includegraphics[width=0.11\textwidth]{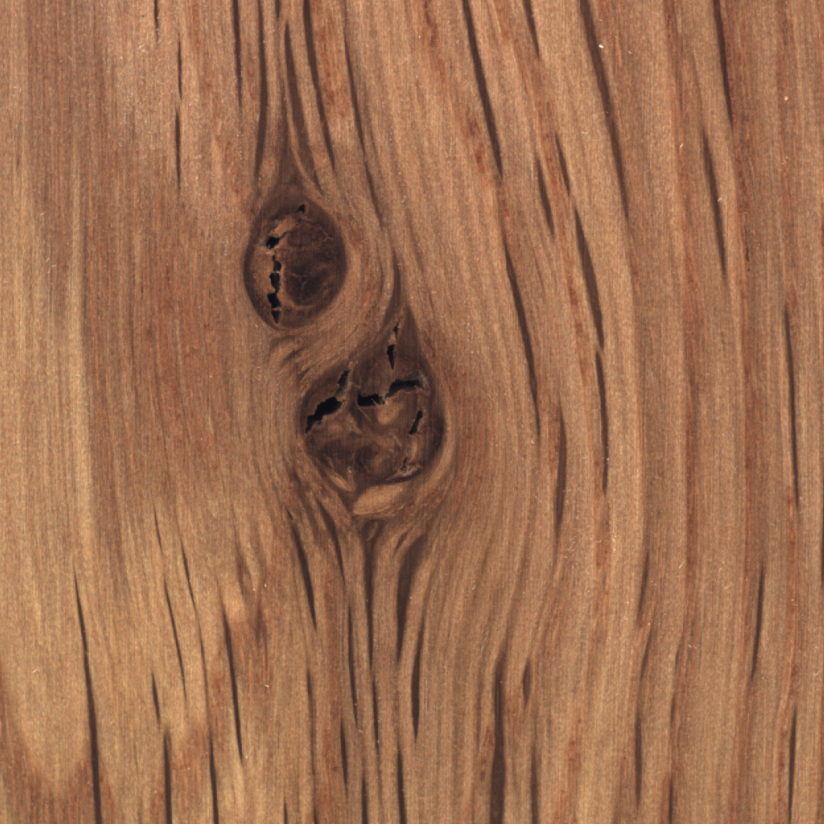} &
\includegraphics[width=0.11\textwidth]{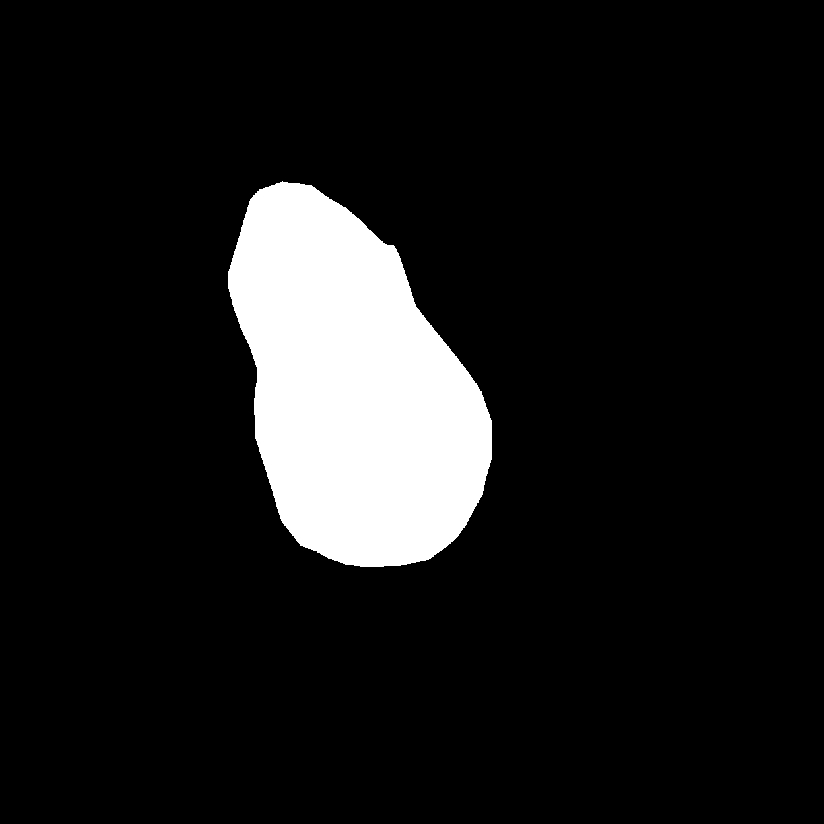} &
\includegraphics[width=0.11\textwidth]{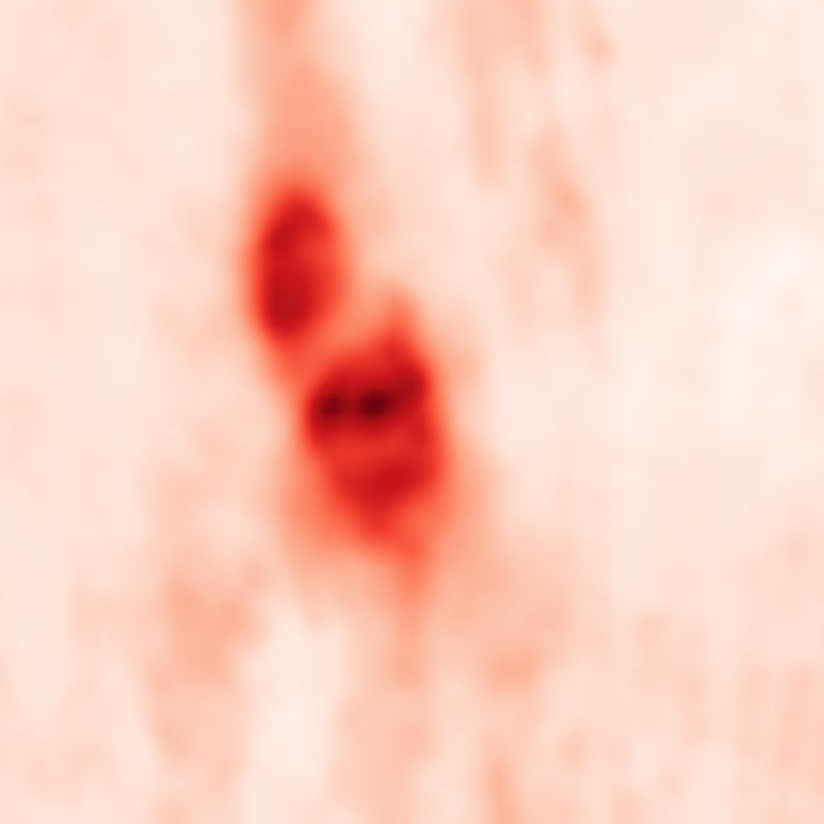} &
\includegraphics[width=0.11\textwidth]{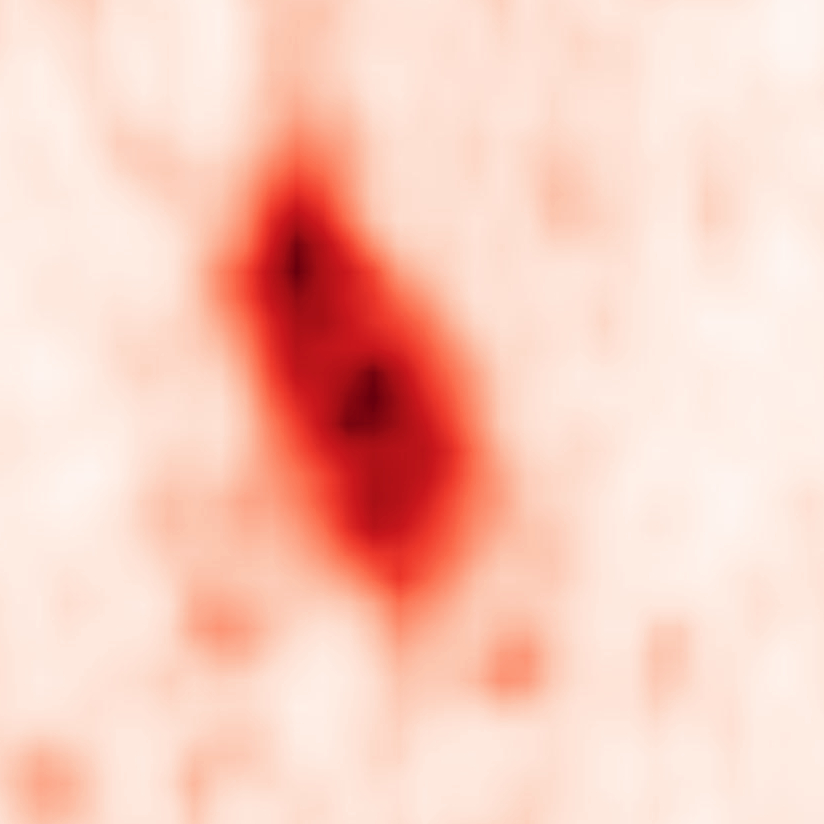} \\
Input Image & Mask & Ours & Aota \etal \cite{aota_zero-shot_2023}
\end{tabular}\\[-0.25\baselineskip]
\caption{\label{fig:gt_check}%
  Manual ground-truth annotations remain subjective where multiple plausible interpretations exist.
  \belowcaptionsqueeze}
\end{figure}

Due to the large and varied datasets, our experiments support the advantage of our method robustly; however, we observe that the manually defined ground-truth masks of MVTec~AD, WFT, and Aitex inevitably introduce a level of subjectivity to those ground-truth references. 
In some cases, as for instance shown in~\Cref{fig:gt_check}, significantly different ground truth interpretations would have been possible, relativizing the accuracy score in such cases.

A limitation of our system is that, by design, it only works on textures. Generic objects can have very different feature statistics in different regions which would not be handled correctly by our method. 

\section{Conclusion}

In this work, we put forward a generic framework for performing zero-shot anomaly localization. 
We identify the importance of the different components and suggest a new approach that significantly improves upon prior art. 
The most important novelty is the proposed FCA for patch statistics comparison which enables high-fidelity anomaly localization that scales well with large textures. 
The performance of the method is validated on several datasets offering a comprehensive overview of the advantages of the method and the trade-off between running time and prediction quality.

\myparagraph{Acknowledgements.}
This project has received funding from the European Union’s Horizon 2020 research and innovation programme under the Marie Skłodowska-Curie grant agreement No 956585.


{\small
\bibliographystyle{ieee_fullname}
\bibliography{ref}
}

\clearpage
\appendix

\twocolumn[
\begin{center}
\Large\textbf{\emph{Supplementary Material }}
\vspace{1cm}
\end{center}
]

\maketitle

\section{Detailed Quantitative Comparison}

\begin{table*}[ht]
\centering

\begin{tabular}{l|c c c|c c c|c c c}
 & \multicolumn{3}{c|}{\textbf{Ours}}  & \multicolumn{3}{c|}{\textbf{Ours + KNN}}  & \multicolumn{3}{c}{\textbf{Aota \etal}\cite{aota_zero-shot_2023}} \\
\hline
\textbf{MVTec AD} \cite{bergmann2019mvtec} & PRO & AUROC & F1 & PRO & AUROC & F1 & PRO & AUROC & F1 \\ 
\hline 
carpet & 95.44 & 98.30 & \textbf{72.58} & \textbf{96.92} & 98.81 & 71.53 & 96.13 & \textbf{98.83} & 69.97 \\ 
grid & \textbf{98.07} & \textbf{99.46} & \textbf{61.62} & 97.77 & 99.27 & 52.41 & 97.01 & 99.12 & 51.89 \\ 
leather & 98.90 & 99.45 & \textbf{66.06} & \textbf{98.92} & \textbf{99.52} & 60.26 & 98.13 & 99.47 & 58.74 \\ 
tile & \textbf{96.33} & \textbf{98.22} & \textbf{82.16} & 88.95 & 94.31 & 65.55 & 84.29 & 93.41 & 63.74 \\ 
wood & \textbf{97.18} & \textbf{98.22} & \textbf{76.34} & 95.22 & 96.98 & 62.42 & 93.56 & 96.54 & 58.68 \\ 
\hline 
\textbf{DTD-Synthetic} \cite{aota_zero-shot_2023} & PRO & AUROC & F1 & PRO & AUROC & F1 & PRO & AUROC & F1 \\ 
\hline 
Blotchy\_099 & 98.67 & 99.55 & 78.92 & \textbf{98.73} & \textbf{99.57} & \textbf{79.50} & 97.60 & 99.19 & 69.56 \\ 
Marbled\_078 & 97.97 & 99.33 & 76.25 & \textbf{98.05} & \textbf{99.37} & \textbf{76.60} & 96.51 & 98.80 & 66.11 \\ 
Mesh\_114 & 94.44 & 97.63 & 65.46 & \textbf{95.91} & \textbf{98.16} & \textbf{66.95} & 95.14 & 97.75 & 64.16 \\ 
Stratified\_154 & 98.78 & 99.15 & 66.67 & \textbf{98.81} & 99.19 & \textbf{66.80} & 98.53 & \textbf{99.25} & 64.48 \\ 
Woven\_068 & 97.24 & 98.86 & 70.51 & \textbf{97.31} & \textbf{98.92} & \textbf{70.72} & 95.34 & 98.29 & 65.61 \\ 
Woven\_125 & 98.51 & 99.52 & 77.03 & \textbf{98.66} & \textbf{99.56} & \textbf{77.59} & 97.00 & 98.99 & 67.50 \\ 
Fibrous\_183 & 96.95 & 98.96 & 72.82 & \textbf{97.21} & \textbf{99.06} & \textbf{73.29} & 94.42 & 98.20 & 65.42 \\ 
Matted\_069 & \textbf{89.43} & 99.33 & 76.15 & 88.89 & \textbf{99.37} & \textbf{76.60} & 89.34 & 99.17 & 68.64 \\ 
Perforated\_037 & 94.55 & 96.60 & 64.66 & \textbf{96.62} & \textbf{97.76} & \textbf{68.78} & 95.74 & 97.05 & 67.41 \\ 
Woven\_001 & 94.73 & 98.93 & 66.17 & \textbf{97.79} & \textbf{99.59} & \textbf{68.96} & 96.51 & 99.42 & 63.59 \\ 
Woven\_104 & 89.98 & 96.84 & 64.93 & \textbf{90.40} & \textbf{96.96} & \textbf{66.05} & 89.78 & 96.60 & 65.28 \\ 
Woven\_127 & 85.26 & 91.63 & 58.88 & \textbf{92.82} & \textbf{94.62} & \textbf{69.58} & 85.96 & 92.22 & 63.81 \\ 
\hline 
\textbf{WFT} \cite{Bergmann2018ImprovingUD} & PRO & AUROC & F1 & PRO & AUROC & F1 & PRO & AUROC & F1 \\ 
\hline 
texture\_1 & \textbf{92.38} & \textbf{97.98} & \textbf{80.34} & 85.36 & 94.25 & 68.46 & 89.01 & 95.94 & 73.12 \\ 
texture\_2 & 86.77 & \textbf{98.56} & \textbf{77.91} & \textbf{87.12} & 98.13 & 77.05 & 80.16 & 96.28 & 71.02 \\ 
\hline 
\textbf{Aitex} \cite{silvestre2019public} & PRO & AUROC & F1 & PRO & AUROC & F1 & PRO & AUROC & F1 \\ 
\hline 
t\_00 & \textbf{78.00} & \textbf{95.08} & 49.41 & 75.19 & 94.08 & \textbf{49.56} & 62.91 & 90.21 & 45.33 \\ 
t\_01 & 76.85 & 91.28 & 69.62 & \textbf{80.08} & \textbf{92.34} & \textbf{73.47} & 70.71 & 91.31 & 71.04 \\ 
t\_02 & 96.34 & \textbf{99.32} & 56.30 & \textbf{96.57} & \textbf{99.33} & \textbf{56.93} & 92.92 & 99.02 & 53.45 \\ 
t\_03 & 88.77 & 97.94 & \textbf{68.04} & \textbf{89.44} & \textbf{97.97} & 67.75 & 87.43 & 97.59 & 65.83 \\ 
t\_04 & \textbf{99.39} & \textbf{99.79} & \textbf{72.44} & \textbf{99.39} & \textbf{99.79} & 72.23 & 97.89 & 99.75 & 72.04 \\ 
t\_05 & \textbf{98.15} & \textbf{99.17} & \textbf{49.34} & 98.04 & 99.11 & 47.66 & 97.95 & 99.05 & 43.41 \\ 
t\_06 & \textbf{99.97} & \textbf{99.99} & 71.60 & 99.97 & \textbf{99.99} & 70.73 & \textbf{99.98} & \textbf{99.99} & \textbf{76.54} \\ 
\hline 

\end{tabular}

\caption{\label{tab:detailed} Metrics breakdown into texture classes.}

\end{table*}

In Table \ref{tab:detailed} we present the detailed breakdown of the metrics into texture classes for all datasets. We report the PRO(0.3), AUROC, and $F_1$ metrics as introduced in the main paper.

\section{Additional Qualitative Comparison}

As mentioned in the main paper, we add here more visualizations for an extensive quality comparison between our method and the current state-of-the-art zero-shot method of Aota \etal~\cite{aota_zero-shot_2023}.

\subsection{Thresholded Results}

\begin{figure*}[ht]
\setlength{\tabcolsep}{1pt}
\renewcommand{\arraystretch}{0.7}
\let\imw\relax
\newlength{\imw}
\setlength{\imw}{0.162\textwidth}
\centering
\begin{tabular}{@{}c@{\hspace{2mm}}c@{\hspace{1mm}}c@{\hspace{1mm}}c@{\hspace{1mm}}c@{\hspace{1mm}}c@{}}

\multirow{2}{*}[3em]{\rotatebox[origin=c]{90}{ MVTec AD}}
  & \includegraphics[width=\imw]{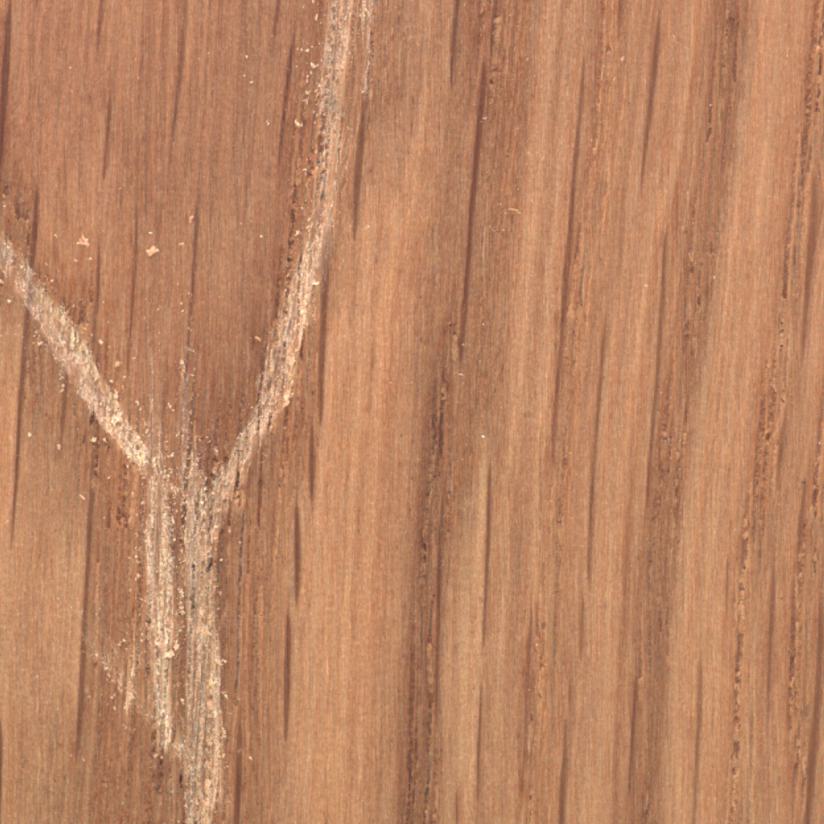} & 
 \includegraphics[width=\imw]{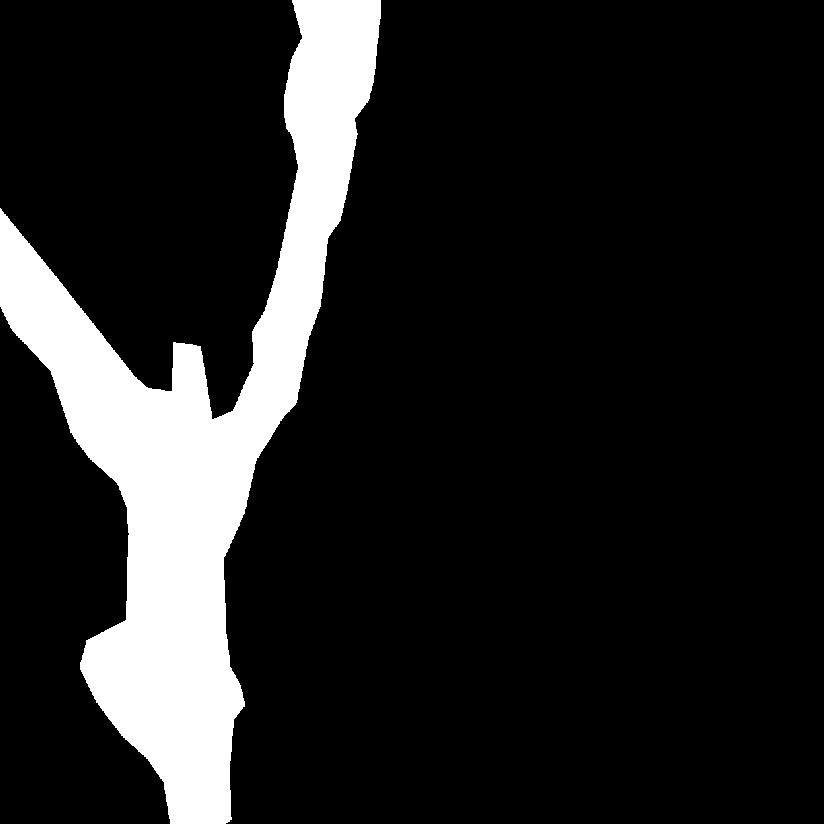} & 
 \includegraphics[width=\imw]{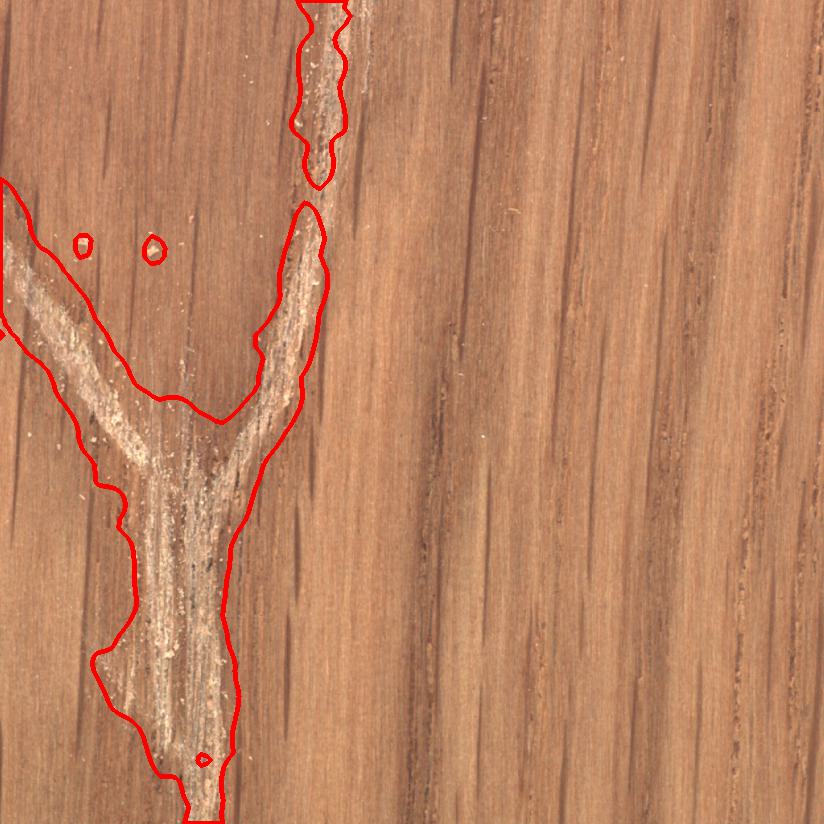} & 
 \includegraphics[width=\imw]{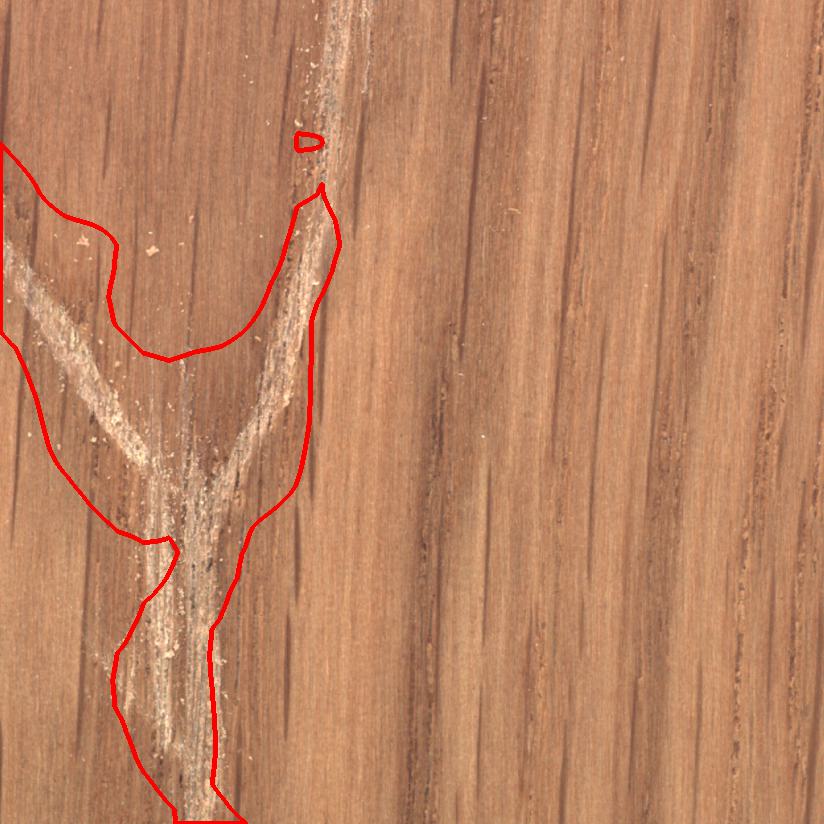} & 
 \includegraphics[width=\imw]{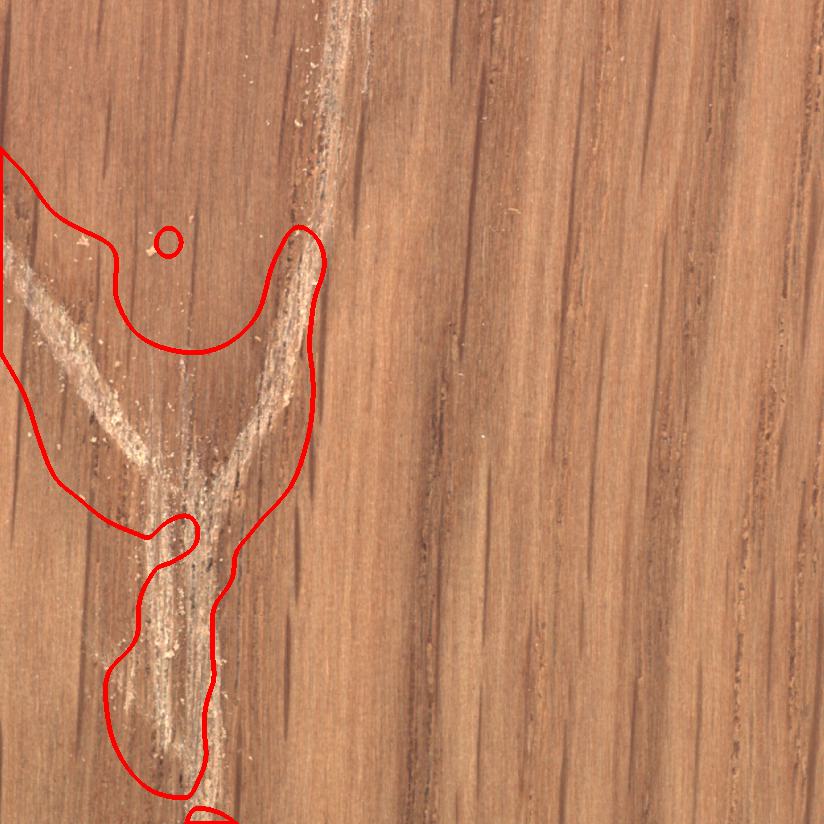} \\ 

 & \includegraphics[width=\imw]{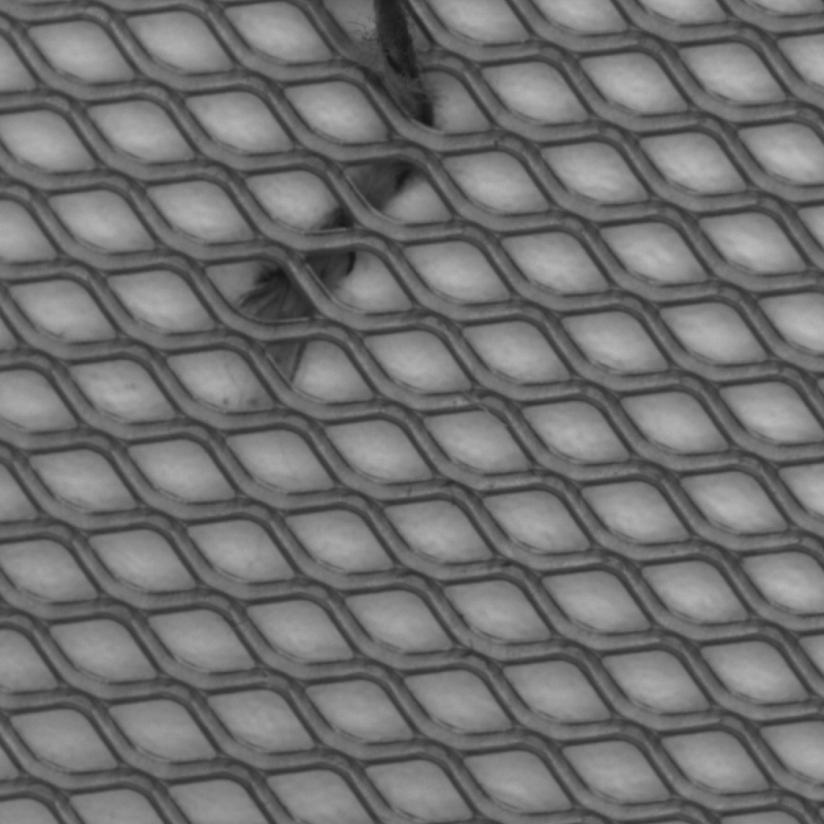} & 
 \includegraphics[width=\imw]{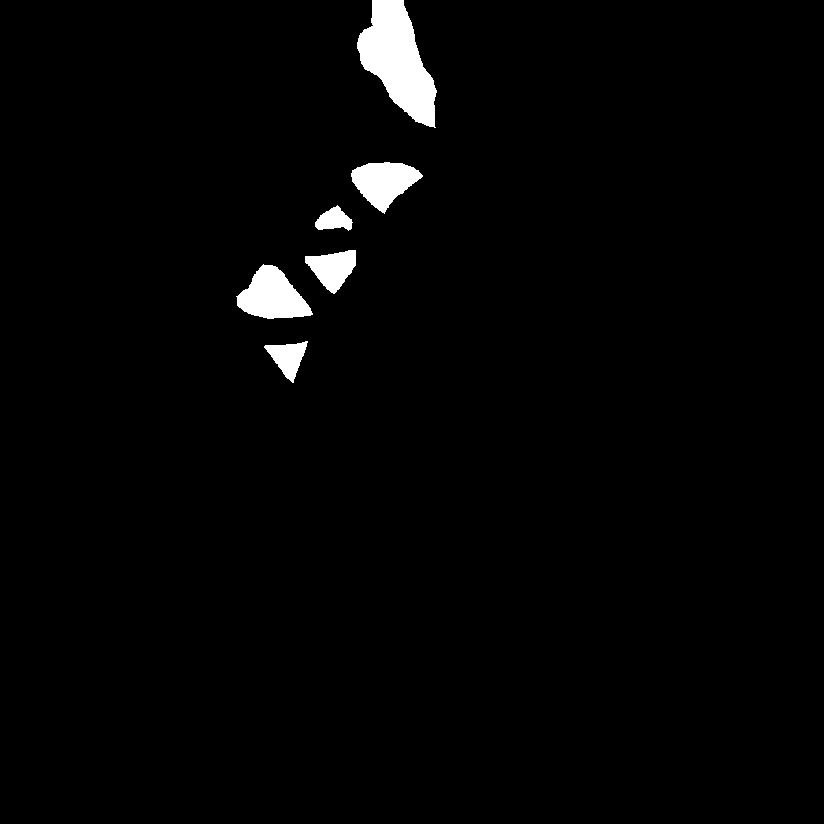} & 
 \includegraphics[width=\imw]{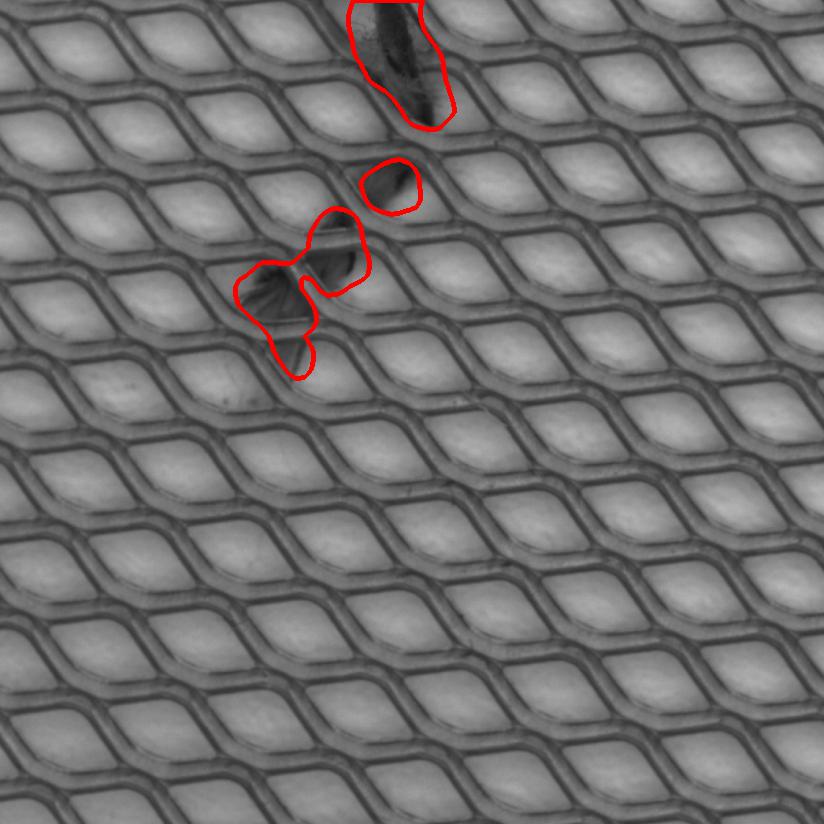} & 
 \includegraphics[width=\imw]{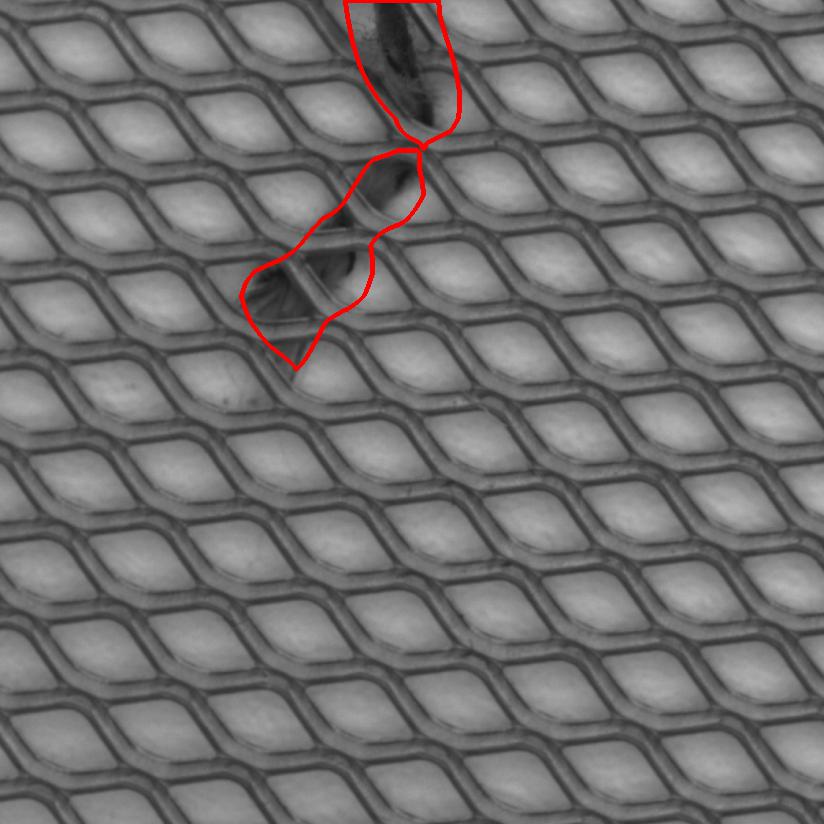} & 
 \includegraphics[width=\imw]{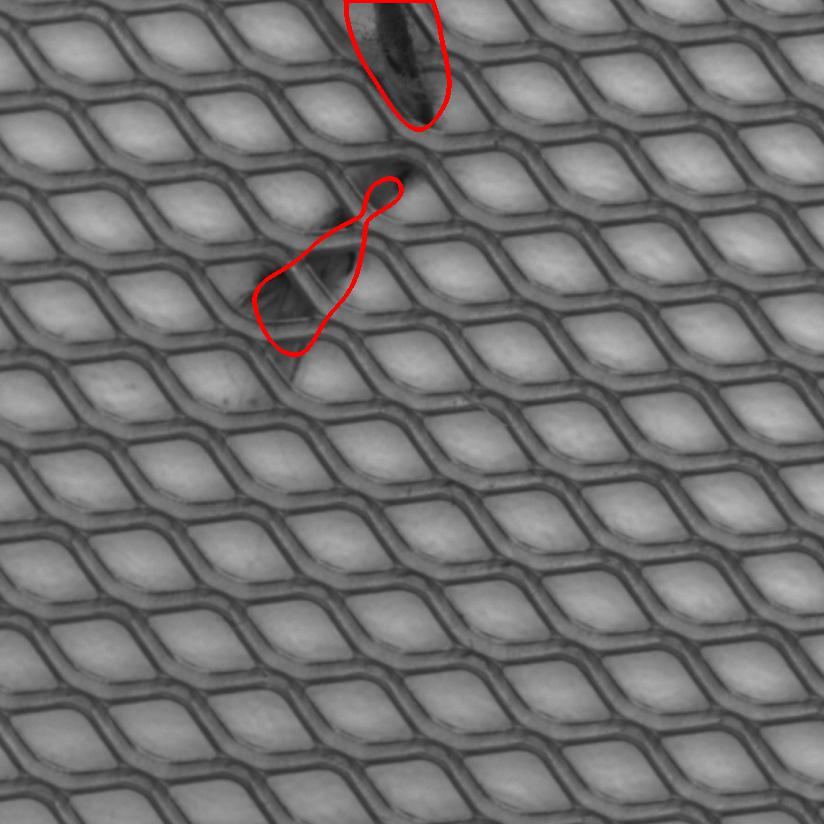} \\ 

\multirow{2}{*}[2em]{\rotatebox[origin=c]{90}{DTD-synthetic}}
& \includegraphics[width=\imw]{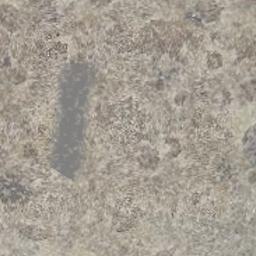} & 
 \includegraphics[width=\imw]{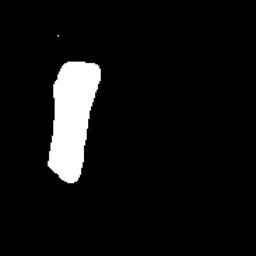} & 
 \includegraphics[width=\imw]{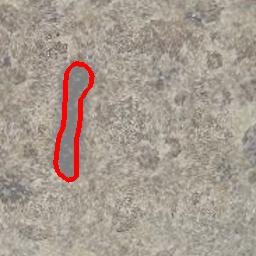} & 
 \includegraphics[width=\imw]{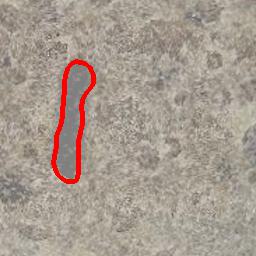} & 
 \includegraphics[width=\imw]{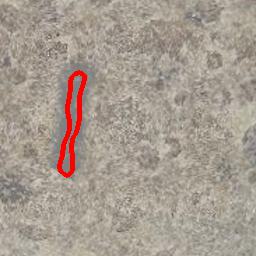} \\ 
 & \includegraphics[width=\imw]{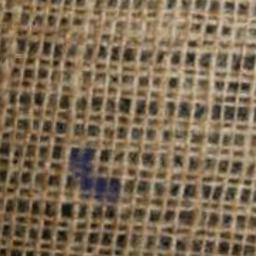} & 
 \includegraphics[width=\imw]{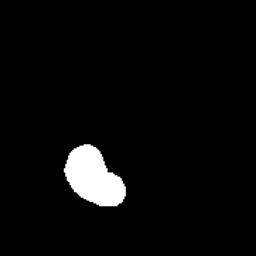} & 
 \includegraphics[width=\imw]{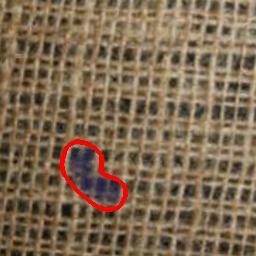} & 
 \includegraphics[width=\imw]{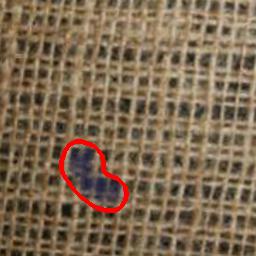} & 
 \includegraphics[width=\imw]{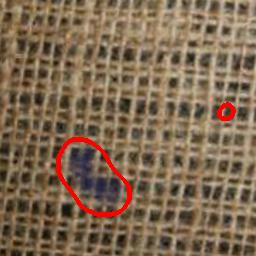} \\ 

\multirow{2}{*}[3em]{\rotatebox[origin=c]{90}{Woven Fabric Textures}}
& \includegraphics[width=\imw]{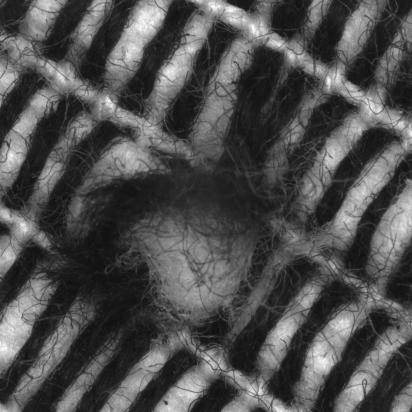} & 
 \includegraphics[width=\imw]{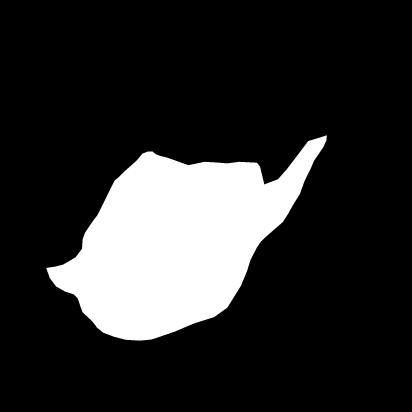} & 
 \includegraphics[width=\imw]{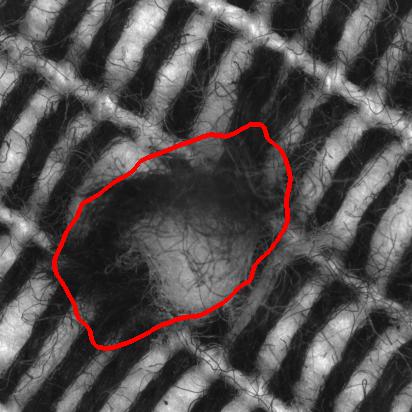} & 
 \includegraphics[width=\imw]{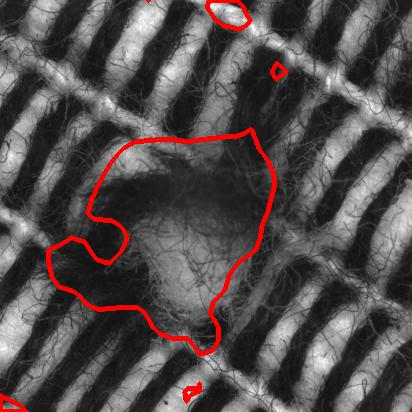} & 
 \includegraphics[width=\imw]{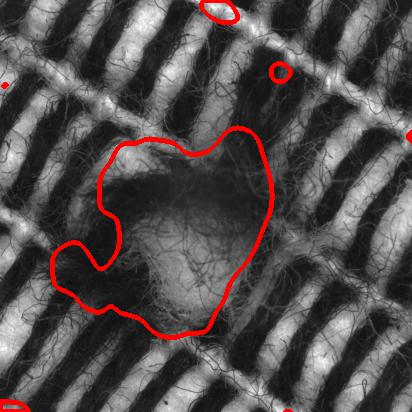} \\ 
 & \includegraphics[width=\imw]{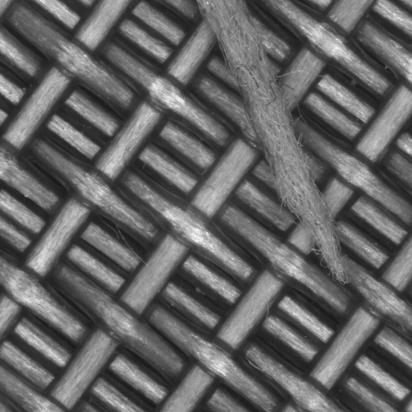} & 
 \includegraphics[width=\imw]{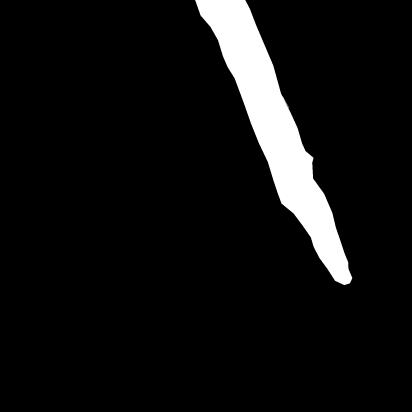} & 
 \includegraphics[width=\imw]{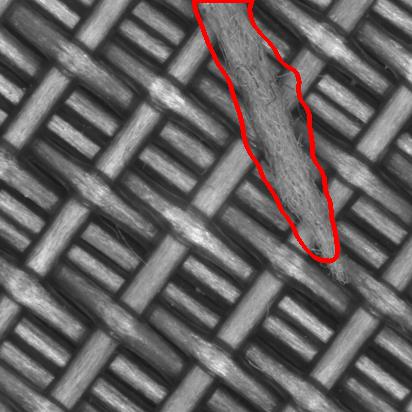} & 
 \includegraphics[width=\imw]{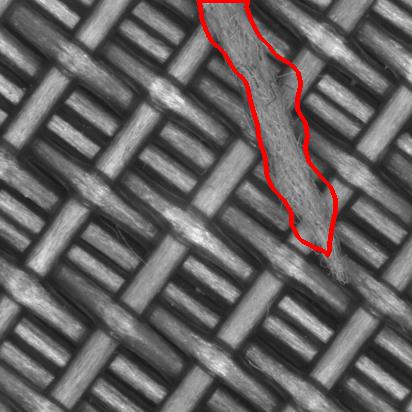} & 
 \includegraphics[width=\imw]{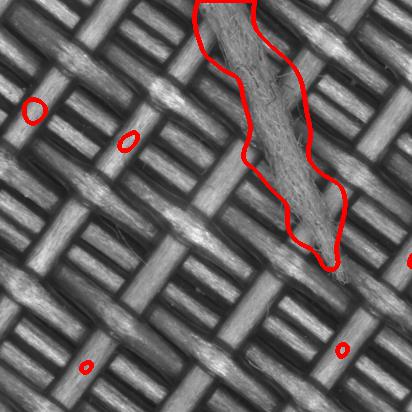} \\ 

\multirow{1}{*}[5em]{\rotatebox[origin=c]{90}{Aitex}}
 & \includegraphics[width=\imw]{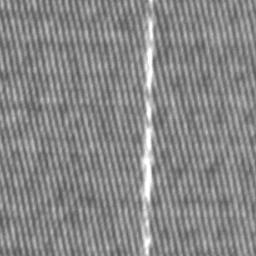} & 
 \includegraphics[width=\imw]{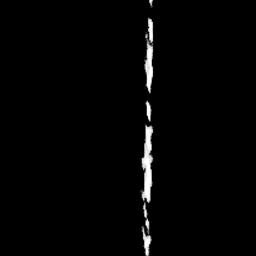} & 
 \includegraphics[width=\imw]{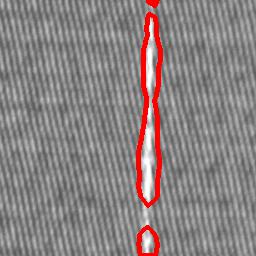} & 
 \includegraphics[width=\imw]{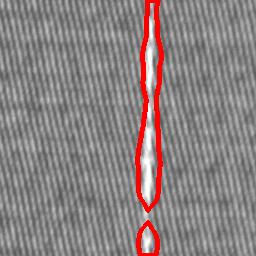} & 
 \includegraphics[width=\imw]{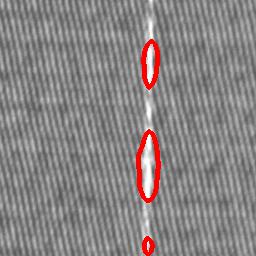} \\ 

& Input Image & GT mask & Ours & Ours$_{320}$ + KNN & Aota \etal \cite{aota_zero-shot_2023}
\end{tabular}
\caption{Qualitative comparison on challenging examples, displaying anomalous regions detected by each method via thresholding of the predicted anomaly maps. The respective thresholds are chosen to be $F_1$-optimal. The thresholded (binary) maps are represented through their enclosing contours; all images are shown after cropping to the center.}
\label{fig:threshold}
\end{figure*}

We visually compare the anomalous regions extracted by different methods, by thresholding the anomaly maps. We compute the optimal threshold with respect to the $F_1$ measure for each texture class individually and display the anomalous area in the original texture. This visualization is included in Figure~\ref{fig:threshold}.

\subsection{Complete Results}

To facilitate the comparison between our method and the main baseline
of Aota \etal \cite{aota_zero-shot_2023}, we present the entirety of
our anomaly maps using a static html tree. The results can be
conveniently browsed under
{\footnotesize\texttt{\href{https://reality.tf.fau.de/pub/ardelean2024highfidelity.html}{reality.tf.fau.de/pub/ardelean2024highfidelity.html}}}.

To get from continuous anomaly scores to the visualizations we perform the following. We consider each texture category/class independently and normalize the scores (same linear scale and bias within each category) for each anomaly prediction, mapping the whole class's minimum and maximum to 0 and 1, respectively. As described in the main paper, the borders ($\approx 10\%$) of each image are ignored during normalization, and they are afterward clipped accordingly. The continuous scores are then mapped to the ``Reds'' color scheme.

\section{Failure Cases}
\begin{figure}[ht]
\setlength{\tabcolsep}{1pt}
\renewcommand{\arraystretch}{0.7}
\setlength{\imw}{0.24\linewidth}
\centering
\begin{tabular}{@{}c@{\hspace{1mm}}c@{\hspace{1mm}}c@{\hspace{1mm}}c@{}}

\includegraphics[width=\imw]{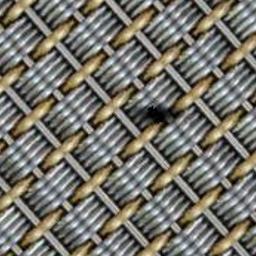} & 
\includegraphics[width=\imw]{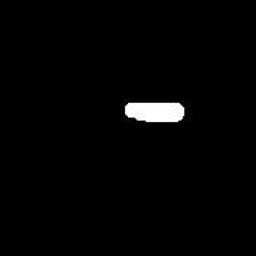} & 
\includegraphics[width=\imw]{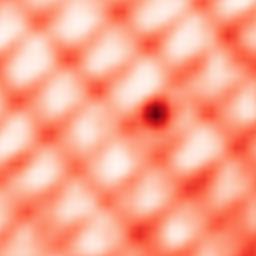} & 
\includegraphics[width=\imw]{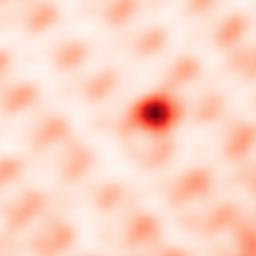} \\

\includegraphics[width=\imw]{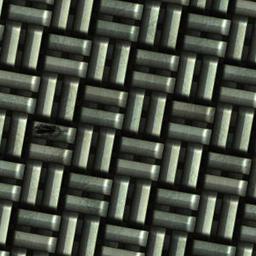} & 
\includegraphics[width=\imw]{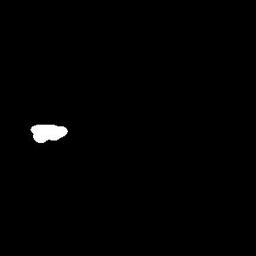} & 
\includegraphics[width=\imw]{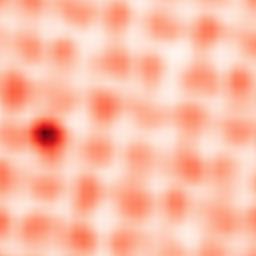} & 
\includegraphics[width=\imw]{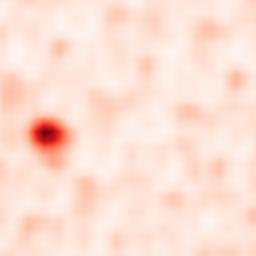} \\

\includegraphics[width=\imw]{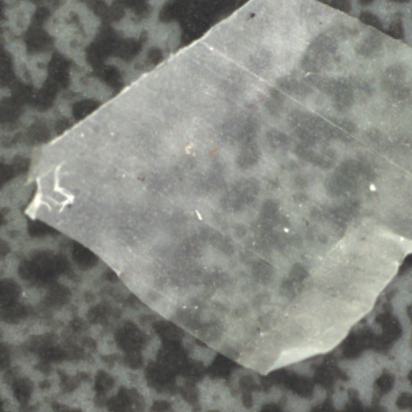} & 
\includegraphics[width=\imw]{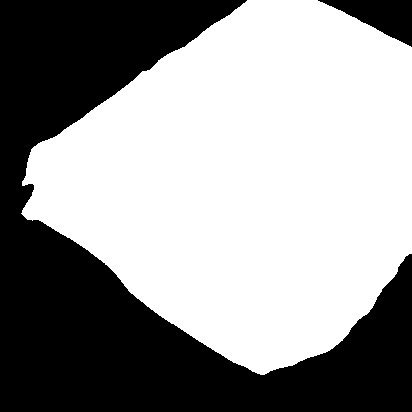} & 
\includegraphics[width=\imw]{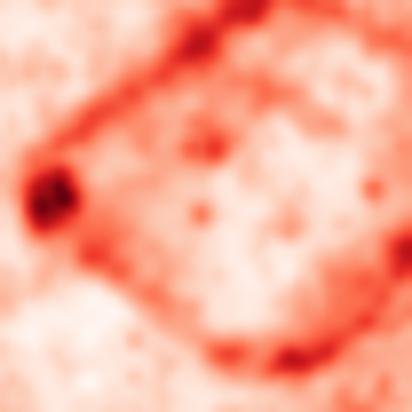} & 
\includegraphics[width=\imw]{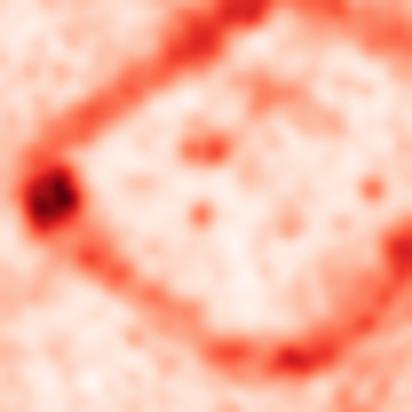} \\

Input Image & GT mask & Ours & Ours + \knn
\end{tabular}
\caption{Visualization of failure cases caused by complex normal distributions with large texture period or anomaly regions that cover the majority of an image. All images are shown after cropping to the center.
}
\label{fig:failure_cases}
\end{figure}

\Cref{fig:failure_cases} investigates the failure cases of our method.
FCA is designed to perform anomaly localization for textured images by comparing patches with a global reference.
As the complexity of the normal texture distribution increases, a single aggregated reference cannot fully capture this complexity.
As shown in the first two rows, complex bimodal textures with large periodicity present difficulties for our method when a single reference is used.
Taking the \knn references alleviates the issue; however, as explained in the main paper, this solution is not feasible for high-resolution images.
The third row in \Cref{fig:failure_cases} shows a different type of failure.
Our formulation assumes that there are enough normal patches in the original image to infer normalcy.
When testing on a center-cropped image from MVTec AD tile class (center crop of $512\times 512$), one can see that if anomalies cover the majority of the image, our method is unable to properly detect these defects.

\section{Additional Study}

We further inspect the proposed approach through a direct ablation and study of the FCA sensitivity to its parameters.

\subsection{Direct Ablation}

In \Cref{tab:sablation}, we perform a direct ablation of the components which constitute the proposed method. 
Ablating \textbf{F} means instead of the default features, extracted by a WideResnet-50~\cite{BMVC2016_87} pretrained on ImageNet, we use the same architecture but no pretraining, \ie randomly initialized weights.
Ablating \textbf{R} means instead of our Wasserstein-optimal feature set (see Section 3.3 main text), we simply take the average of all feature patches in the image.
Ablating \textbf{HR} gives the difference between running at full resolution ($1024 \times 1024$) over low resolution ($320 \times 320$).
Ablation of \textbf{FCA} is performed by replacing our novel Patch Statistics Comparison method with the EMD between histograms of the features in each patch. In this case, the reference is the global histogram.

\begin{table}[ht]
    \centering
    
    \begin{tabular}{c c c c |c c}
    \hline
    F & R & HR & FCA & PRO(0.3) & AUROC \\
    \hline
    \xmark & \cmark & \cmark & \cmark & 70.24 & 87.99 \\
    \cmark & \xmark & \cmark & \cmark & 69.62 & 78.58 \\
    \cmark & \cmark & \xmark & \cmark & 95.46 & 97.74 \\
    \cmark & -- & \cmark & \xmark & 95.35 & 98.00 \\
    \cmark & \cmark & \cmark & \cmark & \textbf{97.18} & \textbf{98.73} \\
    \hline
    \end{tabular}
    \caption{\label{tab:sablation} 
    Ablating different components of our pipeline:
    F - Feature Extractor, R - Reference, HR - Running the method at high resolution.
    See text for more details.
    }
    
\end{table}

\subsection{Sensitivity to parameters}

Our method has only 3 parameters: $T$, $\sigma_p$, and $\sigma_s$.
Additionally, the image size can be considered a preprocessing parameter that influences the performance of the method.
In the main paper, we demonstrated the robustness of our approach to the choice of these parameters by running all experiments with fixed $\sigma_p=3$ and $\sigma_s=1$. We set $T = 9$ when running at full resolution and $T = 3$ at low resolution ($320 \times 320)$. 
This fixed setup performed well on all datasets tested. 

\begin{figure*}

  \includegraphics[width=0.475\linewidth]{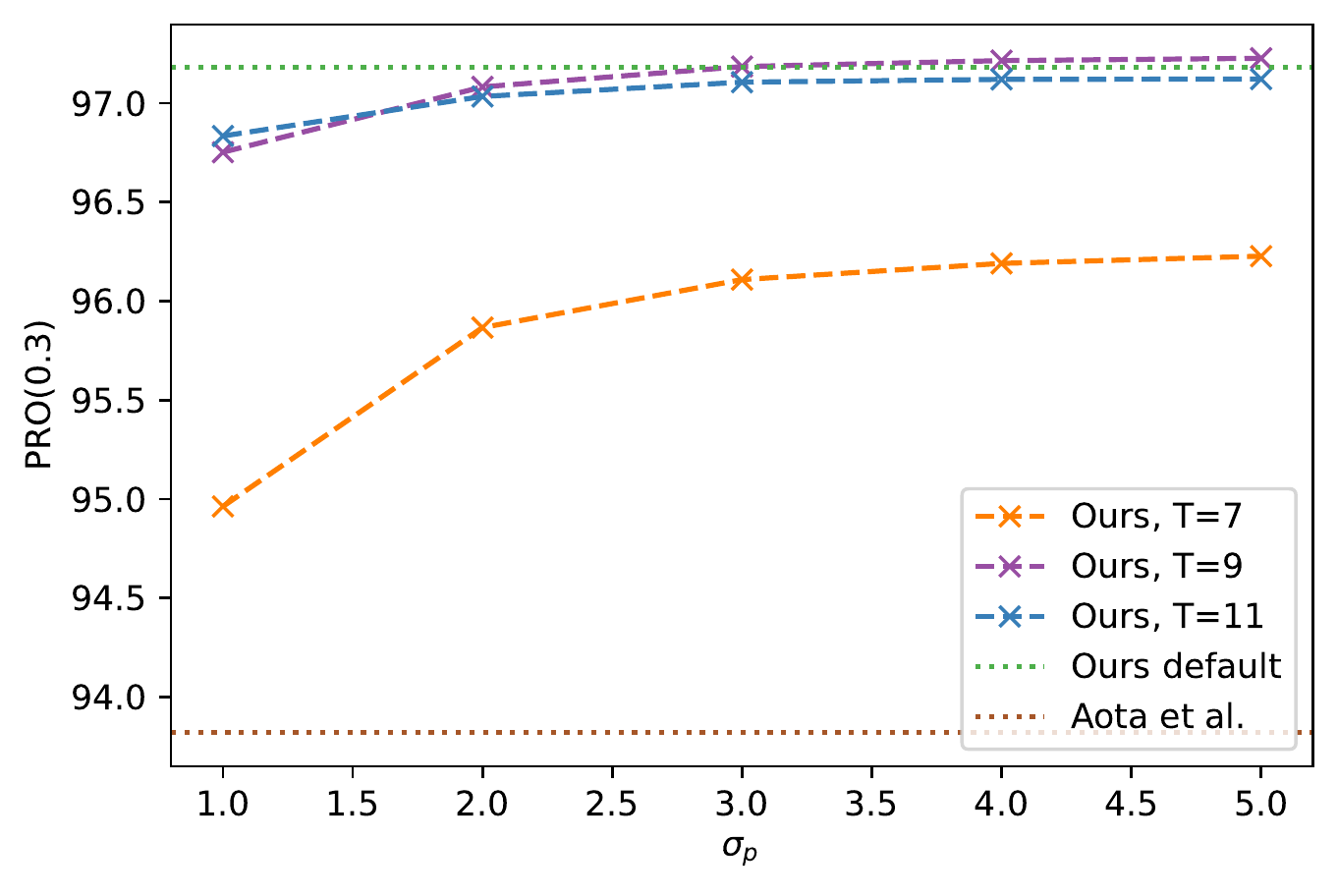}
  \strut\hfill\strut
  \includegraphics[width=0.475\linewidth]{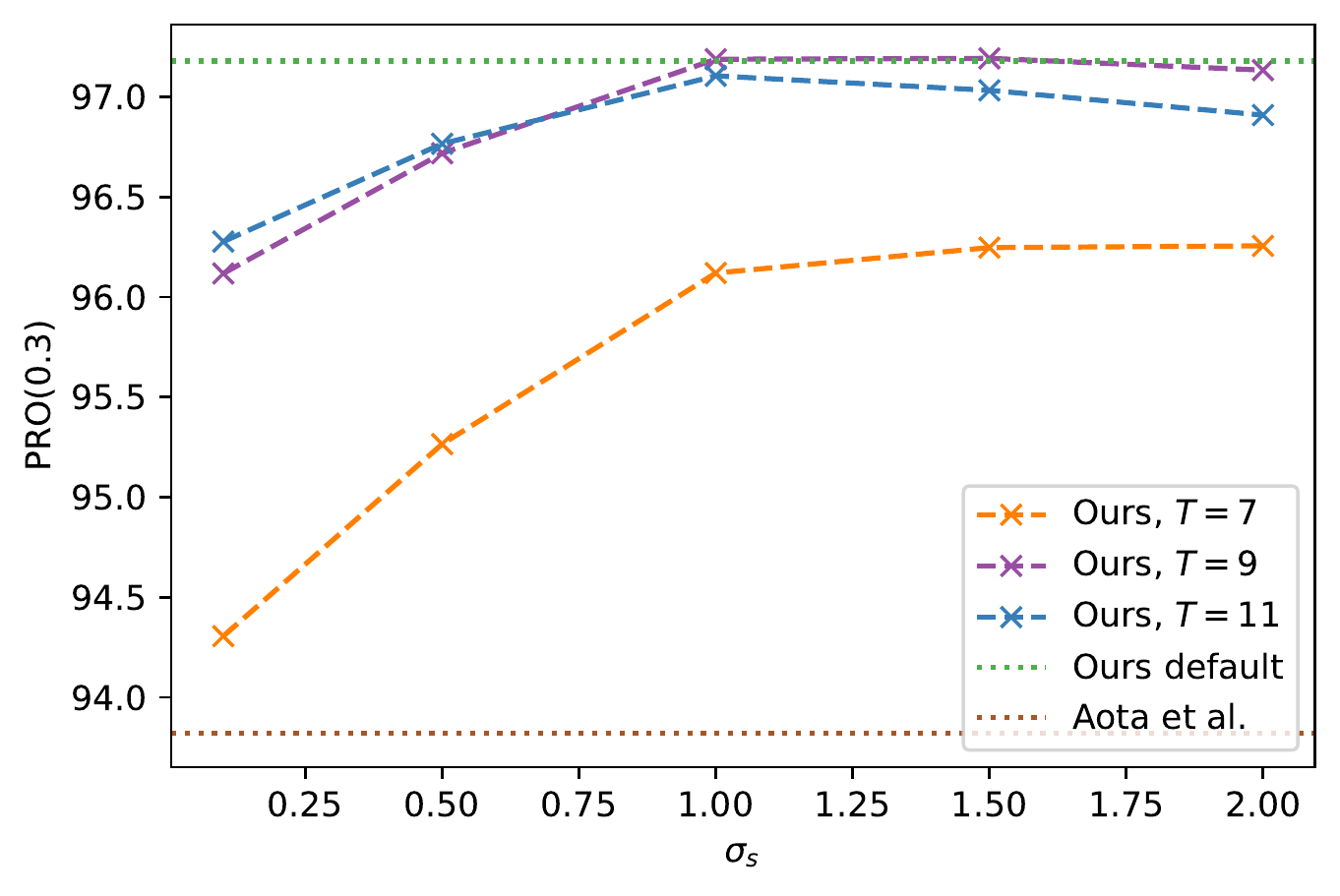}
  \caption{\label{fig:parameters}
    Sensitivity of the method to parameters' variations for the MVTec~AD dataset, at full resolution ($1024\!\times\!1024$). `Ours default' is obtained with $T=9$, $\sigma_p=3.0$, and $\sigma_s=1.0$.
  }
\end{figure*}

In \Cref{fig:parameters} we further show the relationship between varying parameters and the performance on MVTec AD\cite{bergmann2021mvtec}. 
We run our method at full resolution (1024 x 1024) in three settings with $T \in \{7, 9, 11\}$. 
In each case, we vary $\sigma_p$ and $\sigma_s$ and observe that FCA is robust to these variations. 
The performance deteriorates significantly when $T=7$, which is primarily due to a reduced receptive field leading to more false positive predictions.
In general, $T$ should be set large enough to capture the periodicity of the texture.

\section{Miscellaneous Details}

When evaluating the different options for patch statistics comparison (Section 3.2 main text) we use $\sigma=6$ for Gaussian spatial smoothing for all methods.
In the case of FCA, $\sigma_p = 6$ and $\sigma_s = 3$. These hyperparameters are calibrated for the preliminary experiment, where the size of feature maps is significantly larger ($256\times256$).

When computing the histogram method for comparing patches we scale the features to the interval $[0.0, 1.0]$ and use a fixed number of $10$ bins.
Using more bins yields marginal improvements in quality but increased computational cost.
We also scale features to $[0.0, 1.0]$ when computing SWW and FCA to ensure a similar scale across channels and images.

When computing the scores for Aota \etal~\cite{aota_zero-shot_2023}, SAA+~\cite{cao2023segment}, and April-GAN~\cite{chen2023zero} the official code has been used and run with the default parameters; for Saliency~\cite{cheng2014global} we used a public unofficial implementation\footnote{\href{https://github.com/congve1/SaliencyRC}{https://github.com/congve1/SaliencyRC}}; for Bellini \etal~\cite{bellini2016time} we reimplemented the method described by the authors; for PatchCore~\cite{roth_towards_2022}, RD++~\cite{Tien_2023_CVPR}, MAEDAY~\cite{schwartz_maeday_2022}, and WinCLIP~\cite{Jeong_2023_CVPR} we take the results directly from the papers which introduce the respective methods.

\section{Code}

The source code for the introduced method is available at {\footnotesize\texttt{\href{https://github.com/TArdelean/AnomalyLocalizationFCA}{github.com/TArdelean/AnomalyLocalizationFCA}}}.

\end{document}